\documentclass[10pt,journal]{IEEEtran}
\usepackage{graphicx}
\usepackage{amsmath}
\usepackage{amsfonts}
\usepackage{amssymb}
\usepackage[amssymb]{SIunits}
\usepackage{longtable}%flexible table definitions

\usepackage{import}%to allow import of svg drawings
\usepackage{color}%to import svg drawings in color
\usepackage{amsthm}
\usepackage{cite}
\usepackage{subcaption}
\usepackage{scalerel,amssymb}

\usepackage{amssymb}
\usepackage{algorithmic}
\usepackage{algorithm}
\usepackage{rotating}

\usepackage{lipsum}% http://ctan.org/pkg/lipsum
\usepackage{multicol}% http://ctan.org/pkg/multicols
\usepackage{multirow}
\usepackage{url}
\usepackage{enumitem}

\ifCLASSINFOpdf
\else
\fi
\newtheorem{mydef}{Definition}
\newtheorem{myremark}{Remark}

\newtheorem{myprop}{Proposition}

\begin{document}
%\title{SSFN: Low Complexity Self Size-estimating Feed-forward Neural Network using \\ Layer-wise Convex Optimization}
%\title{Self Size-estimating Feed-forward Network, Consistent Monte-Carlo Performance \\ and Low Complexity Design with \\ Limited Human Intervention}

\title{SSFN -- Self Size-estimating Feed-forward Network with Low Complexity, Limited Need for Human Intervention, and Consistent Behaviour across Trials}


\author{\IEEEauthorblockN{Saikat Chatterjee\IEEEauthorrefmark{1},~\IEEEmembership{Member,~IEEE,} Alireza M. Javid\IEEEauthorrefmark{1},  Mostafa Sadeghi\IEEEauthorrefmark{2}, Shumpei Kikuta\IEEEauthorrefmark{3}, Dong Liu\IEEEauthorrefmark{1}, \\ Partha P. Mitra\IEEEauthorrefmark{4}, Mikael Skoglund\IEEEauthorrefmark{1},~\IEEEmembership{Fellow,~IEEE}} \\
	\IEEEauthorblockA{\IEEEauthorrefmark{1} School of Electrical Engineering \& Computer Science, KTH Royal Institute of Technology, Sweden} \\
	\IEEEauthorblockA{\IEEEauthorrefmark{2} INRIA Grenoble Rhone-Alpes, France} \\
	\IEEEauthorblockA{\IEEEauthorrefmark{3} University of Tokyo, Tokyo, Japan} \\
	\IEEEauthorblockA{\IEEEauthorrefmark{4} Cold Spring Harbor Laboratory, 1 Bungtown Road, New York, USA}}  % <-this % stops a space

\IEEEtitleabstractindextext{%
\begin{abstract}
We design a self size-estimating feed-forward network (SSFN) using a joint optimization approach for estimation of number of layers, number of nodes and learning of weight matrices. The learning algorithm has a low computational complexity, preferably within few minutes using a laptop. In addition the algorithm has a limited need for human intervention to tune parameters. SSFN grows from a small-size network to a large-size network, guaranteeing a monotonically non-increasing cost with addition of nodes and layers. The learning approach uses judicious a combination of `lossless flow property' of some  activation functions, convex optimization and instance of random matrix. Consistent performance - low variation across Monte-Carlo trials - is found for inference performance (classification accuracy) and estimation of network size.

\end{abstract}

}

\maketitle

% is not selected <OR> if conference mode is selected - because compsoc
% conference papers position the abstract like regular (non-compsoc)
\IEEEdisplaynontitleabstractindextext

% \fi
\IEEEpeerreviewmaketitle

%\ifCLASSOPTIONcompsoc
%\IEEEraisesectionheading{\section{Introduction}\label{sec:introduction}}
%\else

\section{Introduction}
\label{sec:Introduction}

Artificial neural networks (ANNs) are popular for pattern classification applications \cite{Connectionist_Handwriting_Recognition_2009,Representation_Learning_Review_2013,3D_CNN_2013,Feedforward_LSTM_LanguageModeling_2015}. Feedforward neural network is a common ANN architecture that continues to attract researchers' attention. Let us assume that a feed-forward neural network \cite{Bebis_FeedForwardNeuralNet_1994} has $L$ layers, and its $l$'th layer has $n_l$ nodes. Estimation of the number of layers $L$ and the number of nodes $\{ n_l \}_{l=1}^L$ helps to decide the size of the network for achieving good performance \cite{SizeMattersInANN_1999}. Estimation of $L$ and $\{ n_l \}_{l=1}^L$ is a combinatorial optimization problem. A significant human intervention is in vogue to address the optimization problem using extensive trial-and-error, often driven by experience, intuition and hand tuning. Our interest is to develop a computationally simple algorithmic solution that involves a limited need for human intervention. 

In pursuit of an algorithmic solution, we address a joint optimization approach to estimate the numbers $L$ and $\{ n_l \}_{l=1}^L$, and learn weight matrices. We start with a small-size feed-forward neural network, and add new nodes and layers, resulting in a large-size structure (wide and deep). Eventually our proposed algorithm decides $L$ and $\{ n_l \}_{l=1}^L$, and learns the weight matrices. We refer to the neural network as self size-estimating feed-forward network (SSFN). 
%We mention that SSFN is a suboptimal solution of the original combinatorial problem of estimating $L$ and $\{ n_l \}_{l=1}^L$, and non-convex problem of learning weight matrices. 

The algorithm for increase in size of SSFN from a small-size neural network to a large-size network ensures a monotonically non-increasing training cost. In the algorithm, we add a new layer on top of an existing structure and optimize parameters for the newly added layer. This is a layer-wise optimization approach. Optimization at each layer seeks an optimal estimate of the target for that layer, ensuring the monotonically non-increasing training cost with increase in size.

Increase in size of SSFN leads to increase in number of parameters and eventually overfitting to the training dataset. Regularization helps to address overfitting. Often parameters of regularization techniques are hand-tuned.
%Norm-based regularization is a standard approach to address overfitting, for example, use of $\ell_2$ or $\ell_1$ norm of weight matrices. 
We address regularization of weight matrices of SSFN analytically, without hand-tuning. Our layer-wise optimization approach allows to establish analytical forms for the regularization parameters. 
%An analytical form for a regularization parameter is advantageous as it helps to avoid tedious cross-validations. 
To find the analytical forms, we use a property of a single-layer feed-forward neural network (SLFN) system. The property ensures that the output of the SLFN system is exactly equal to the input to the SLFN system. That means an input signal flows through the SLFN without any loss or change. We refer to the property as `lossless flow property' (LFP). We provide sufficient conditions to construct an SLFN such that it holds LFP. Using the LFP, we find analytical forms of regularization parameters for multi-layer SSFN.
%and ensure monotonic decrease in the training cost with increase in size of SSFN.

Further, the layer-wise non-convex optimization is relaxed to a convex problem for an efficient use of regularization. This relaxation requires to use a structure in weight matrices. 
%Every weight matrix becomes rank deficient. 
For a weight matrix, a part is learned and the other part is chosen as an instance of a random matrix. Use of random matrix instance in weight matrix construction brings questions on consistent behaviour across independent simulations / trials. Consistent behaviour across independent simulations is an important aspect of a reliable solution. We use layer-wise sequential optimization and hence a random matrix instance chosen for the weight matrix of the first layer affects buildup of next layers. Similarly random matrix instances of all preceding layers affect succeeding layers. To study consistent behaviour, we observe how the size of SSFN varies across Monte-Carlo simulations as well as how the inference performance varies. Use of Monte-Carlo simulations closely corresponds to a situation where several researchers are independently looking for reproducibility. Achieving consistent behaviour with a limited human intervention is a challenging problem for neural networks, including deep learning methods. 

In our experiments, we use  eight popular benchmark datasets for sound and image classification tasks. For each of the eight datasets that we experiment with, we will observe that size and performance both have low variations across Monte-Carlo simulations. On the other hand, size of SSFN varies significantly across the eight datasets. Finally, in contrast to success of SSFN for several datasets, we show a failure case. The SSFN provides a significantly poor performance for CIFAR-10 dataset compared to the state-of-the-art \cite{Cubuk_AutoAugment_2018}. This case illustrates a limitation of SSFN. To mitigate the limitation, we develop a simple ad-hoc approach. 
We use the well-known alternating-direction-method-of-multipliers (ADMM) \cite{BoydPCPE11} to realize the layer-wise convex optimization. The use of convex optimization and the further use of ADMM lead to a significantly low computational complexity requirement. The computational complexity of SSFN learning algorithm is in order of  minutes when executed in a standard laptop for the eight datasets. Finally we study the use of backpropagation for further improvement of weight matrices in SSFN at the expense of more computation. We mention that the idea of SSFN was first shown in \cite{PLN_Saikat} as the name `progressive learning network' and subsequently improved in this article with appropriate theoretical supports and extensive experimental evaluations. We decided not to use the name `progressive learning network' in this article to avoid confusion as the same name is used for other schemes in literature, that are not relevant.

\subsection{Literature Survey}
There exists a vast literature on neural network design.
% and theoretical underpinning on function approximation capabilities \cite{Hopfield_ANN_1988, FUNAHASHI_ANN_1989, HORNIK_UniversalApproximation_1991}. Function approximation analyzes a mapping function between the input signal of a neural network and the targeted output. 
%Currently we did not pursue theoretical analysis for function approximation capability of PLN.
%Structure of neural network, in particular the size of the network, matters in practice \cite{SizeMattersInANN_1999}.
Training of multi-layer neural network has received a significant attention. 
%A standard anastz is that a gradient-based optimization starting from random initialization may get stuck near poor locally optimum solutions. 
An approach to constructive addition of layers and use of supervised learning was explored in  \cite{Cascade-correlation_ANN_1990,Training_MLPs_layer_by_layer_1996}. Recently, deep learning structures (with many layers) \cite{Representation_Learning_Review_2013,lbdl15,blda09,GoodBC16} have attracted a high attention in literature. Additionally, structures in weight matrices such as convolutional neural networks \cite{Hinton_DeepCNN_2012}, structure of connection between layers such as residual networks \cite{DeepResidualLearning_CVPR_2016}, and structures based on feedback such as recurrent neural networks \cite{RNN_based_language_model_2012} have been explored. Deep belief network (DBN) \cite{DBN_Hinton_2006} and its variants \cite{Bengio_GreedyLayerWise_DNN_2007} use greedy layer-wise unsupervised learning for creating an initial network and then, further training using backpropagation for supervised learning. 
Examples of existing greedy and/or layer-wise learning approaches can be found in  \cite{Ivakhnenko_Polynomial_Theory_of_Complex_Systems_1971,blglw07,KulkK17,HettCEHJW17}. Then, examples of advanced regularization methods and practical approaches, such as softweights, dropout, can be found in \cite{Larsen_regularized_ANN_1994, Hinton_SoftWeight_1992, Hinton_Dropout_2014}. 
%There are new works in advanced methods of regularization, for example, use of data augmentation \cite{Cubuk_AutoAugment_2018}, domain knowledge based dropout in input signals as well as model parameters \cite{DeVries_Cutout_2017}, etc. 
In the backdrop of above-mentioned works, the SSFN is a feed-forward neural network. It uses supervised learning to minimize a cost for the training dataset while estimating its own size. The regularization coefficients are analytically derived to minimize the cost. 

%Every weight matrix in SSFN has a rank deficient structure. Reduction in number of parameters in a deep learning structure can be handled by parsimonious models, such as low-rank weight matrices \cite{LowRankMatrixInANN_2013}. In SSFN, a part of the rank-deficient weight matrix is optimized and the other part is a random matrix instance.
For a layer of SSFN, a part of the weight matrix is optimized and the other part is an instance of random matrix.
% There exists ample precedence in machine learning and neural networks for use of random matrix instance. Random instance based initialization is common in unsupervised learning, for example, K-means algorithm. Neural networks also use random matrix based initialization and then optimize using back propagation. 
% There are works in neural networks to use random matrix instances for weight matrices that are not optimized. A relevant example is extreme learning machine (ELM) \cite{HuanZS06,HuanZS04,HuanZDZ12} where weight matrices for all the layers except the last layer (or extreme layer) are typically random matrix instances. The weight matrix for the last layer of ELM is optimized. 
There are works in the neural network literature that use random matrix instances for weight matrices. Prominent examples are extreme learning machines  \cite{HuanZS06,Huang_What_are_ELM_2015,Huang_ELM_2012}. There are several other neural networks with random matrix based weights, discussed in the survey article \cite{Review_ANN_random_weights_2018}. 
% where weight matrices for all the layers except the last layer (or extreme layer) are typically random matrix instances. The weight matrix for the last layer of ELM is optimized. 
%For ELM, there exist theoretical arguments supporting the use of random instances  \cite{Huang_What_are_ELM_2015,Huang_UniversalApproximation_ELM_2006}. In practice, implementation of ELM is simple and an ELM shows good performance for several applications. We compare PLN with ELM in our object recognition experiments. 
%, while ELM structure is chosen a-priori.
%Related methods based on random matrix instances in neural networks and then, further extension to kernel methods such as random kitchen sinks are in \cite{Schimdt_NeuralNetWithRandomWeights_1992, PAO_NN_RandomVectors_1994, Igelnik_FunctionApproximation_StochasticChoice_1995, LU_ANN_RandomWeights_2014, Cao_NeuralNetWithRandomWeights_2015, Rahimi_RandomKitchenSinks_NIPS_2008, FastFood_AlexSmola_2013}. 
Related methods based on random matrix instances in neural networks and then, further extension to kernel methods such as random kitchen sinks are in \cite{Schimdt_NeuralNetWithRandomWeights_1992, PAO_NN_RandomVectors_1994, Igelnik_FunctionApproximation_StochasticChoice_1995, LU_ANN_RandomWeights_2014, Rahimi_RandomKitchenSinks_NIPS_2008, FastFood_AlexSmola_2013}. 
The major difference with prior works is that our proposed SSFN has partially optimized weight matrices for all the layers. In addition SSFN estimates its own size. 
At this point, we mention that the use of random matrix instances is well accepted in signal processing and information theory, for example, in compressed sensing (CS) \cite{Donoho_2006_Compressed_sensing,CS_introduction_Candes_Wakin_2008,Vehkapera_Kabashima_Chatterjee_TIT_2016}. %In CS, a small-dimensional feature vector is collected from a large dimensional sparse vector. 
A closely related field to CS is sparse representation and dictionary learning, successfully used for face recognition and image classification \cite{Sastry_2009_Face_recognition, Label_Consistent_KSVD_2013}. There are endeavors to connect iterative sparse representation (sparse recovery) algorithms and multi-layer neural networks by algorithms unrolling where iterations are viewed as layers \cite{Learning_fast_approximations_of_sparse_coding_LeCun_2010,Learned_Convolutional_Sparse_Coding_2018,Algorithm_Unrolling_YCEldar_2019}.

%Finally we mention that our proposed work is different than the `progressive learning machine' of \cite{Progressive_Learning_Machine_ANN_2015} that addresses hybrid system. Our proposed PLN does not address hybrid system. The PLN is also different from `progressive learning technique' of \cite{Progressive_learning_technique_ANN_2016} that addresses adaptation to new classes in a multi-class classification problem. The PLN does not consider adaptation to new classes. 

A relevant topic area is neural network architecture search (NAS), where diverse methods have been applied. Example approaches are based on evolutionary algorithms \cite{Todd_1988,Miller_Todd_1989,Kitano_1990}, reinforcement learning \cite{NAS_RL_2017} and Bayesian learning \cite{pmlr-v64-mendoza_towards_2016,Bayesian_NAS_2019}. Recent works on NAS with several references can be found in the survey article \cite{NAS_Survey_2018}. 
%Many of the recent works require a high level of computational complexity, for example,  the work of \cite{NAS_RL_2017} used 800 GPUs for three to four weeks.
Many of the NAS works require a high level of computational resource, for example, the work of \cite{NAS_RL_2017} used 800 GPUs. The survey article \cite{NAS_Survey_2018} repeatedly mentions requirement of high computational requirement in the order of many GPU days.
Further, many NAS works have a common aspect that architecture search and training for optimization of parameters are separated. 
%The work \cite{Bayesian_NAS_2019} addressed architecture search and training of weight matrices together in a variational Bayes principle. 
The `future directions' section of the survey article \cite{NAS_Survey_2018} mentions about the complexity for a fair comparison of diffrent NAS methods and reproducibility of published results. It is argued that performance of a NAS method depends on many factors other than the architecture itself. The factors can be search space, computational budget, data augmentation (ex. CutOut, MixUp), training procedures, regularization (ex. Dropout, Shake-shake), etc. It is therefore conceivable that these factors may have a significant impact on reported performance numbers than the better architectures found by NAS. Instead of many factors we concentrate on designing a feedforward network as a core architecture. We rely on commonly used signal processing and optimization tools, and look for a low complexity solution, with limited human intervention, that provides consistent and reproducible behaviour. We also mention a failure case in contrary to a typical apprehension towards showing negative results.

%In the backdrop of NAS methods, the proposed SSFN is a convex optimization based low complexity solution for the joint optimization of estimating the size of a feed-forward network architecture and learning the weight matrices. SSFN requires order of minutes in a standard laptop. We are also interested to design the SSFN without much human intervention in parameter tuning. Therefore we rely on commonly used signal processing and optimization tools, and look for consistent and reproducible behaviour. 
%We also report the failure case for CIFAR-10 dataset where the SSFN provides poor performance in a consistent manner.

%\subsection{Organization of article and notations}
%Organization of this article is follows. We provide literature survey below. Progression property is introduced in Section**. Proposed PLN is discussed in Section**. Finally experiments are shown in Section** followed by conclusions at last.

\section{Design of SSFN}
In this section we engineer the proposed self size-estimating feed-forward network (SSFN) and provide some theoretical underpinning. 
%SSFN is a feed-forward neural network. 
%Basic block diagram of a feed-forward neural network is shown in Figure~\ref{fig:Multi_layer_ANN}. 
We begin with the original optimization problem in the next subsection and then develop SSFN in the following subsections. 

%\begin{figure}[t!]
%	\centering
%	\def\svgwidth{\linewidth}
%	\input{fig_I_1.eps_tex}	
%	\caption{Structure of a feed-forward neural network. The top diagram is a conventional illustration where a `violet' color circle represents an activation node and a `blue' color circle represents a scalar in estimated target output vector. The bottom diagram is for our own illustration. LT stands for \emph{linear transform} (weight matrix) and NLT stands for \emph{non-linear transform} (activation function). Input data/signal is denoted by $\mathbf{x}$ and output is  $\tilde{\mathbf{t}}$. The signal transformation at $l$'th layer is denoted by $\mathbf{y}_l$. In literature, the vector $\mathbf{y}_l$ is commonly known as feature vector for the $l$'th layer. }
%	\label{fig:Multi_layer_ANN}
%\end{figure}

\subsection{Optimization problem}
In a supervised learning problem, let $(\mathbf{x},\mathbf{t})$ be a pair-wise form of the data vector $\mathbf{x}\in \mathbb{R}^P$ that we observe, and the target vector $\mathbf{t} \in \mathbb{R}^Q$ that we wish to infer. The target vector $\mathbf{t}$ can be a categorical variable for a classification problem with $Q$-classes.
%Typically $Q < P$, but that need not be always. 
Let us construct a feed-forward neural network with $L$ layers, and $n_l$ nodes in the $l$'th layer. We denote the weight matrix for $l$'th layer by $\mathbf{W}_l \in \mathbb{R}^{n_{l} \times n_{l-1}}$. For an input vector $\mathbf{x}$, a feed-forward neural network produces a transformation $\mathbf{f}: \mathbb{R}^P \rightarrow \mathbb{R}^{n_L}$ in its last layer. The transformation depends on parameters as
\begin{eqnarray}
\mathbf{f} \triangleq \mathbf{f} \left(\mathbf{x}, L, \{ n_l \}_{l=1}^L, \{ \mathbf{W}_l \}_{l=1}^L \right).
\end{eqnarray} 
Then, we use a linear transformation to generate the target prediction $\tilde{\mathbf{t}} = \mathbf{Of}$ where $\mathbf{O} \in \mathbb{R}^{Q \times n_L}$ is the output system matrix. We assume that there exists no parameter to optimize for activation functions; activation functions are well-defined and fixed. 

The training phase of the neural network considers estimation of parameters $L, \{ n_l \}$, and learning of parameters $\{ \mathbf{W}_l \}$ and $\mathbf{O}$. Suppose that we have a $J$-sample training dataset $\mathcal{D} = \{ (\mathbf{x}^{(j)},\mathbf{t}^{(j)} ) \}_{j=1}^J$. We define the cost function 
\begin{eqnarray}
\begin{array}{rl}
\mathcal{C} \!\!\!\! &= \! \frac{1}{J} \sum_{j=1}^J \| \mathbf{t}^{(j)}  - \tilde{\mathbf{t}}^{(j)} \|^2 \\
& =  \! \frac{1}{J} \sum_{j=1}^J \| \mathbf{t}^{(j)} - \mathbf{O} \mathbf{f} \left( \mathbf{x}^{(j)}, \! L, \! \{ n_l \}_{l=1}^L, \! \{ \mathbf{W}_l \}_{l=1}^L \right) \! \|^2. 
\end{array}
\end{eqnarray}
Throughout the article, we use $\| . \|$ to denote $\ell_2$-norm.
The optimization problem is
\begin{eqnarray}
\underset{L, \{ n_l \}, \{ \mathbf{W}_l \}, \mathbf{O}}{\arg\min} \,\, \mathcal{C} \,\,\,\, \mathrm{subject \,\, to} \,\,
\left\{
\begin{array}{l}
L \leq L_{max}, \\
\forall l, n_{min} \leq n_l \leq n_{max}, \\
\forall l, \| \mathbf{W}_l \|_F^2 \leq \nu, \\
\| \mathbf{O} \|_F^2 \leq \epsilon,
\end{array}
\right.
\label{eq:Original_Optimization_Problem}
\end{eqnarray}
where $\|.\|_F$ denotes Frobenius norm.
Here, the constraint $\| \mathbf{W}_l \|_F^2 \leq \nu$ acts as a regularization to avoid overfitting of weight matrices to the training dataset. Similarly, we have the regularization parameter $\epsilon$ for learning the $\mathbf{O}$ matrix. Assume that we have a maximum number of layers allowed, denoted by $L_{max}$. Similarly, we have a minimum and a maximum number of nodes in every layer, denoted by $n_{min}$ and $n_{max}$, respectively. We have two challenging aspects, discussed below.
\begin{enumerate}
	\item The optimization problem \eqref{eq:Original_Optimization_Problem} is not only non-convex, but also combinatorial. We have a combinatorial  search problem with exponential complexity $(n_{max}-n_{min})^{L_{max}}$ to choose $L$ and $\{ n_l \}_{l=1}^L$.  It is difficult to find a globally optimum solution of \eqref{eq:Original_Optimization_Problem}. Instead, a good principle for a sub-optimal approach is valuable.
	\item Selection of regularization coefficients, such as $\nu$ and $\epsilon$, is non-trivial. The selection is often addressed by cross-validation, a tedious approach. Instead of cross-validation, an analytical approach is valuable.  
\end{enumerate}

\subsection{Layer-wise non-convex optimization}
\label{subsec:Layer-wise_learning}

In a feed-forward neural network, signal flows in one direction from the input side to the output side. The signal flow relation between $(l-1)$'th layer and $l$'th layer is
\begin{eqnarray}
\mathbf{y}_l = \mathbf{g}(\mathbf{W}_l \, \mathbf{y}_{l-1}) \in \mathbb{R}^{n_l}, \,\, l=1,2, \hdots ,L,
\label{eq:SignalFlow_in_FeedforwardANN}
\end{eqnarray} 
where $\mathbf{y}_l$ denotes signal transformation at the $l$'th layer; for the first layer $\mathbf{y}_0 = \mathbf{x}$. In literature, the vector $\mathbf{y}_l$ is commonly known as feature vector for the $l$'th layer.
Let $g(.)$ denotes a non-linear activation function such as rectified-linear-unit (ReLU), and $\mathbf{g}(.)$ denotes scalar-wise use of $g(.)$. That means $\mathbf{g}$ function is a stack of $n_l$ nodes where each node uses the activation function $g(.)$ on the corresponding scalar of the $\mathbf{W}_l \, \mathbf{y}_{l-1}$ vector.
A natural question is what will be a good principle for designing a sub-optimal approach to address \eqref{eq:Original_Optimization_Problem} while maintaining the feed-forward signal flow relation  \eqref{eq:SignalFlow_in_FeedforwardANN}.

We have mentioned the need for a good principle to design a sub-optimal approach in the previous subsection. In search of a good principle, we use a layer-wise optimization principle that ensures a monotonically non-increasing cost with the increase in size of SSFN. The principle helps to design an algorithm for estimating size of SSFN and learning parameters with appropriate regularization. We add layers one-by-one in this principle. When we have added a new layer, we add (activation) nodes one-by-one to increase size of the added new layer. This sequential addition allows us to design an automatic algorithm for SSFN construction without much involvement of hand tuning.

For SSFN, addition of a layer is more complex than addition of a node. We first discuss addition of a layer with necessary constraints. 
Suppose that we have an $(l-1)$-layer SSFN that is ready to use. We now add a new layer to construct an $l$-layer SSFN where $l \leq L_{max}$ and with the signal flow relation~\eqref{eq:SignalFlow_in_FeedforwardANN}. The parameters to design the $l$'th layer are $n_l$ and $\mathbf{W}_l$. We use $\mathcal{C}_l$ to denote the cost for the $l$-layer SSFN. The cost $\mathcal{C}_l$ is 
\begin{eqnarray}
\begin{array}{rl}
\mathcal{C}_l \!\!\!\! &= \! \frac{1}{J} \sum_{j=1}^J \| \mathbf{t}^{(j)}  - \tilde{\mathbf{t}}_l^{(j)} \|^2 \\
& =  \! \frac{1}{J} \sum_{j=1}^J \| \mathbf{t}^{(j)} - \mathbf{O}_l \, \mathbf{y}_l^{(j)}  \|^2 \\
& = \! \frac{1}{J} \sum_{j=1}^J \| \mathbf{t}^{(j)} - \mathbf{O}_l \, \mathbf{g}(\mathbf{W}_l \, \mathbf{y}_{l-1}^{(j)})  \|^2,
\end{array}
\label{eq:LayerWise_CostFunction}
\end{eqnarray}
where $\tilde{\mathbf{t}}_l$ denotes the output of the $l$-layer SSFN.
Here $\mathbf{O}_l \in \mathbb{R}^{Q \times n_l}$ is an output system matrix to project the feature vector $\mathbf{y}_l$ to the target vector $\mathbf{t}$. 

In the proposed layer-wise learning principle, we address the following optimization problem for each layer $l = 1,2,\hdots, L, L \leq L_{max}$, starting with $l=1$:
\begin{eqnarray}
\underset{ n_l, \mathbf{W}_l , \mathbf{O}_l}{\arg\min} \,\, \mathcal{C}_l \,\,\,\, \mathrm{subject \,\, to} \,\,
\left\{
\begin{array}{l}
n_{min} \leq n_l \leq n_{max}, \\
\| \mathbf{W}_l \|_F^2 \leq \nu, \\
\| \mathbf{O}_l \|_F^2 \leq \epsilon, \\
\mathcal{C}_l \leq \mathcal{C}_{l-1}^{\star}.
\end{array}
\right.
\label{eq:LayerWise_Optimization_Problem}
\end{eqnarray}
Here, we use the notation $\star$ to represent an optimal value; $\mathcal{C}_{l}^{\star}$ denotes the optimal cost as a result of the above optimization problem. The fourth constraint $\mathcal{C}_l \leq \mathcal{C}_{l-1}^{\star}$ ensures a monotonically non-increasing cost; the constraint ensures $\mathcal{C}_l^{\star} \leq \mathcal{C}_{l-1}^{\star}$ after optimizing~\eqref{eq:LayerWise_Optimization_Problem}. Overall the above optimization problem \eqref{eq:LayerWise_Optimization_Problem} is a sub-optimal approach to address the original optimization problem \eqref{eq:Original_Optimization_Problem} in a sequential manner. 

Considering \eqref{eq:LayerWise_CostFunction}, the optimization problem \eqref{eq:LayerWise_Optimization_Problem} is associated with a single-layer feed-forward network (SLFN) with the input $\mathbf{y}_{l-1}$ and the output $\tilde{\mathbf{t}}_l = \mathbf{O}_l \, \mathbf{g}(\mathbf{W}_l \, \mathbf{y}_{l-1})$. For a given $n_l$, optimization of $\mathbf{O}_l$ and $\mathbf{W}_l$  is non-convex. On the other hand,  finding an appropriate $n_l$ is no more combinatorial. We can start with a preset (minimum) value of $n_l$ and increase it one-by-one or in a step size until the cost minimization shows a saturation trend. 

For further progress with the optimization problem~\eqref{eq:LayerWise_Optimization_Problem}, we now raise two theoretical questions, as follows. 
\begin{enumerate}[noitemsep,nolistsep]
	\item Is the non-convex optimization problem~\eqref{eq:LayerWise_Optimization_Problem} feasible? 
	\item How do we analytically set regularization parameters in ~\eqref{eq:LayerWise_Optimization_Problem} so that we can avoid cross-validation? 
\end{enumerate}
If the optimization problem turns out to be feasible then we raise a practical question: how do we construct the SLFN $\tilde{\mathbf{t}}_l = \mathbf{O}_l \, \mathbf{g}(\mathbf{W}_l \, \mathbf{y}_{l-1})$ with a low complexity?
%If the optimization problem turns out to be feasible, then we raise practical questions as follows.
%\begin{enumerate}
%\item How do we construct the SLFN $\tilde{\mathbf{t}}_l = \mathbf{O}_l \, \mathbf{g}(\mathbf{W}_l \, \mathbf{y}_{l-1})$ with a low complexity?
%\item Can we have a trade-off between computational complexity and model complexity?
%\end{enumerate}
%  That means how do we get a good solution of the optimization problem~\eqref{eq:LayerWise_Optimization_Problem} in practice?
The above mentioned theoretical and practical questions are non-trivial. We address the questions using a specific structure in weight matrix $\mathbf{W}_l$. The structure for weight matrix $\mathbf{W}_l$ is decided by using a property associated with an SLFN. The property is introduced and explained in the next subsection.

\subsection{Lossless flow property (LFP)}
Lossless flow property (LFP) is associated with an SLFN. We will now construct an SLFN that fulfils the LFP. Let us use two variables $\check{\mathbf{t}} \in \mathbb{R}^Q$ and $\hat{\mathbf{t}} \in \mathbb{R}^Q$ to denote the input and output of an SLFN, respectively. The SLFN signal flow relation is  $\hat{\mathbf{t}} = \mathbf{B} \, \mathbf{g}(\mathbf{A} \check{\mathbf{t}}) \in \mathbb{R}^Q$, where $\mathbf{A} \in \mathbb{R}^{n' \times Q}$ is the input-side weight matrix and $\mathbf{B} \in \mathbb{R}^{Q \times n'}$ is the output-side matrix. The number of nodes in the SLFN is denoted by $n'$.

\begin{mydef}[Lossless flow property (LFP)]
	The SLFN fulfils LFP if there are matrices $\mathbf{A},\mathbf{B}$ and appropriate non-linear activation function $g(.)$ such that 
	\begin{eqnarray}
	\hat{\mathbf{t}} = \mathbf{B} \, \mathbf{g}(\mathbf{A} \check{\mathbf{t}}) =  \check{\mathbf{t}}, \forall \check{\mathbf{t}} \in \mathbb{R}^Q.
	\end{eqnarray}
\end{mydef}
\noindent This means that the input $\check{\mathbf{t}}$ to the SLFN flows to the output $\hat{\mathbf{t}}$ without any loss, resulting in $\hat{\mathbf{t}} = \check{\mathbf{t}}$. The $\mathbf{g}(.)$ function has no linear activation function. Each node of the hidden layer of SLFN has the non-linear activation function $g(.)$. Then the question is how to construct an SLFN that holds LFP. We provide the following proposition as sufficient conditions.

\begin{myprop}[LFP holding SLFN]
	\label{prop:ReLU_based_SLFN}
	An SLFN holds LFP if the following conditions hold.
	\begin{enumerate}
		\item We use ReLU activation function, defined as $g(t) = \mathrm{max}(0,t)$, $t \in \mathbb{R}$.
		\item Number of nodes $n' = 2m$ where $m\geq Q$ denotes a new integer variable.
		\item The matrices $\mathbf{A}$ and $\mathbf{B}$ have following factorized structures as 
		$\mathbf{A}=\mathbf{V}_m \mathbf{C}$ and $\mathbf{B} = \mathbf{C}^{\dag} \mathbf{U}_m$, where $\mathbf{C} \in \mathbb{R}^{m \times Q}$ is a full column-rank matrix and $\dag$ denotes pseudoinverse. Here, $\mathbf{V}_m$ and $\mathbf{U}_m$ are two deterministic matrices as follows 
		\begin{eqnarray}
		\mathbf{V}_m  = \left[  
		\begin{array}{c}
		\mathbf{I}_m \\
		- \mathbf{I}_m
		\end{array}
		\right] \,\, \mathrm{and} \,\,\, 
		\mathbf{U}_m = \left[  
		\mathbf{I}_m  \,\,\ - \mathbf{I}_m
		\right],
		\end{eqnarray}
		where $\mathbf{I}_m$ is m-dimensional identity matrix. 
	\end{enumerate} 
	The LFP holding SLFN structure is $\mathbf{C}^{\dag} \mathbf{U}_m \, \mathbf{g}(\mathbf{V}_m \mathbf{C} \check{\mathbf{t}})$ satisfying the above conditions.
\end{myprop}

\noindent Proof: This proposition is a sufficient condition for LFP by existence. Let $\pmb{\gamma} = \mathbf{C} \check{\mathbf{t}} \in \mathbb{R}^m$ where $\mathbf{C}$ is a full column-rank matrix. Use of ReLU activation function results in $\pmb{\gamma} = \mathbf{U}_m \, \mathbf{g}(\mathbf{V}_m \pmb{\gamma})$ and the SLFN output is $\mathbf{C}^{\dag} \pmb{\gamma} = \mathbf{C}^{\dag} \mathbf{C} \check{\mathbf{t}} = \check{\mathbf{t}}$. \hfill $\blacksquare$

We can construct a full-column rank matrix $\mathbf{C}$ in several ways. Examples of $\mathbf{C}$ matrix construction are as follows.
\begin{enumerate}
	\item Using random matrix instance: Generate a random matrix where components are drawn from iid distributions (such as Gaussian or uniform). The matrix is full column-rank with high probability as $m \geq Q$. 
	%Then we can use Gram-schimdt orthogonalization followed by $\ell_2$-norm normalization of columns.  
	\item Using deterministic matrix instance: We can use columns from discrete cosine transform (DCT), Wavelet transform, etc, and their combinations. 
	\item Using a low number of parameters: 
	%We can form a full column-rank matrix using a low number of parameters. 
	For example, we can use toeplitz or circulant structures for creating full-rank square matrix. Circulant structure is associated with a convolutional filter. We can also form a full column-rank matrix from Koronecker product of two small full column-rank matrices.  
	\item A trivial example is the identity matrix, that is $\mathbf{C}=\mathbf{I}_Q$.
\end{enumerate}

We now discuss about a few activation functions, mainly some other derivatives of ReLU, for LFP holding SLFN design. The derivatives are leaky ReLU and a generalized ReLU. The definition of leaky ReLU \cite{MaasHN13} is 
\begin{eqnarray}
g(t) = \left\{ 
\begin{array}{c}
t, \,\, \mathrm{if} \,\, t \geq 0 \\
at, \,\, \mathrm{if} \,\, t < 0,
\end{array}
\right.
\end{eqnarray}
where $0 < a < 1$ is a fixed scalar and typically small. Leaky ReLU based SLFN holds LFP if the conditions in Proposition~\ref{prop:ReLU_based_SLFN} hold with a small modification that $\mathbf{U}_m \triangleq \frac{1}{1+a}\left[ \mathbf{I}_m  \,\, - \mathbf{I}_m \right]$. Generalizing the definition of leaky ReLU, we now define a generalized ReLU function as follows 
\begin{eqnarray}
g(t) = \left\{ 
\begin{array}{c}
bt, \,\, \mathrm{if} \,\, t \geq 0 \\
at, \,\, \mathrm{if} \,\, t < 0.
\end{array}
\right.
\end{eqnarray}
where $a,b > 0$ are fixed scalars with the relation $a < b$. The generalized ReLU based SLFN also holds LFP if the conditions in Proposition~\ref{prop:ReLU_based_SLFN} hold with a small modification that $\mathbf{U}_m \triangleq \frac{1}{a+b}\left[ \mathbf{I}_m  \,\, - \mathbf{I}_m \right]$. While we can use leaky ReLU and generalized ReLU for SSFN construction, we continue to use ReLU activation function in this article.

\subsection{Addressing theoretical questions}

We raised two theoretical questions at the end of Section~\ref{subsec:Layer-wise_learning}.
These questions are concerned with a feasibility study and analytical form of regularization parameters.
We now proceed with the knowledge of LFP. Note that the optimization problem \eqref{eq:LayerWise_Optimization_Problem} is addressed layer-wise where we have access to the optimized $(l-1)$'th layer SSFN, and then address optimization of the $l$'th layer. 

Let us consider the $(l-1)$-layer SSFN where the parameters of $(l-1)$'th layer were optimized by solving \eqref{eq:LayerWise_Optimization_Problem} for $\mathcal{C}_{l-1}$. For the optimized $(l-1)$-layer SSFN, we have the optimal output $\tilde{\mathbf{t}}_{l-1}^{\star} = \mathbf{O}_{l-1}^{\star} \, \mathbf{y}_{l-1}$. This $\tilde{\mathbf{t}}_{l-1}^{\star}$ corresponds to the optimal cost $\mathcal{C}_{l-1}^{\star}$. Note that $\| \mathbf{O}_{l-1}^{\star} \|_F^2 \leq \epsilon$.

We now check feasibility of the optimization problem~\eqref{eq:LayerWise_Optimization_Problem} where we have four constraints. How to set the parameter $n_{min}$ and the regularization parameters $\nu,\epsilon$, while satisfying $\mathcal{C}_l \leq \mathcal{C}_{l-1}^{\star}$?  
The feasibility of optimization problem~\eqref{eq:LayerWise_Optimization_Problem} is stated in the following proposition.

\begin{myprop}[Feasibility]
	The optimization problem~\eqref{eq:LayerWise_Optimization_Problem} is feasible under the following conditions
	\begin{eqnarray}
	\begin{array}{l}
	n_{min} = 2m \geq 2Q, \\ 
	\epsilon = \| \mathbf{U}_m \|_F^2 \, Q = 2mQ, \\
	\nu = 2mQ\epsilon = (2mQ)^2, \,\, \mathit{and} 
	\end{array}
	\end{eqnarray}
	the used activation functions are helpful to fulfil LFP. 
\end{myprop}
\noindent Proof: This proposition is a sufficient condition. We prove the feasibility of~\eqref{eq:LayerWise_Optimization_Problem} by providing an example of feasible solution where $\mathcal{C}_l = \mathcal{C}_{l-1}^{\star}$. There may be many locally optimum solutions of \eqref{eq:LayerWise_Optimization_Problem} for which $\mathcal{C}_l < \mathcal{C}_{l-1}^{\star}$. The feasible solution example is: $\mathbf{W}_l = \mathbf{V}_m \mathbf{C} \mathbf{O}_{l-1}^{\star}$ and $\mathbf{O}_l = \mathbf{C}^{\top} \mathbf{U}_m$, where $\mathbf{C}$ is a full column rank  and orthonormal matrix.
Full column rank requires $m \geq Q$.  Orthonormality satisfies $\mathbf{C}^{\top} \mathbf{C} = \mathbf{I}_Q$, $\mathbf{C}^{\dag} = \mathbf{C}^{\top}$ and $\| \mathbf{C} \|_F^2 = \| \mathbf{C}^{\top} \|_F^2 = Q$.
For this feasible solution to hold, we require $\| \mathbf{O}_l \|_F^2 \leq \| \mathbf{U}_m \|_F^2 \| \mathbf{C}^{\top} \|_F^2 = 2mQ$ and hence, we set $\epsilon =  2mQ$. Also we have $\| \mathbf{W}_l \|_F^2 \leq \| \mathbf{V}_m \|_F^2 \| \mathbf{C} \|_F^2 \| \mathbf{O}_{l-1}^{\star} \|_F^2 = 2mQ\epsilon = (2mQ)^2$ and hence we set $\nu =  (2mQ)^2$.
The feasible solution ensures $\mathcal{C}_l = \mathcal{C}_{l-1}^{\star}$ due to the following relation
\begin{eqnarray}
\begin{array}{rl}
\tilde{\mathbf{t}}_l & =  \mathbf{O}_l \, \mathbf{g}(\mathbf{W}_l \, \mathbf{y}_{l-1}) \\
& = \mathbf{C}^{\top} \mathbf{U}_m \, \mathbf{g}(\mathbf{V}_m \mathbf{C} \mathbf{O}_{l-1}^{\star} \, \mathbf{y}_{l-1}) \\
& =  \mathbf{C}^{\top} \mathbf{U}_m \, \mathbf{g}(\mathbf{V}_m \mathbf{C} \, \tilde{\mathbf{t}}_{l-1}^{\star}) \\
& =  \mathbf{C}^{\dag} \mathbf{U}_m \, \mathbf{g}(\mathbf{V}_m \mathbf{C} \, \tilde{\mathbf{t}}_{l-1}^{\star}) \\
& =  \tilde{\mathbf{t}}_{l-1}^{\star}.
\end{array}
\end{eqnarray}
In the above derivation, we use LFP in the last step. A sufficient condition for LFP to hold is $m \geq Q$. For the feasible solution, we require $n_l =  2m \geq 2Q$. Hence we have $n_{min} = 2m \geq 2Q$. \hfill $\blacksquare$

The above proposition provides a set of analytically driven choice of regularization parameters. In addition, it provides a suggestion on the minimum number of nodes per layer and how to set $\mathbf{W}_l$ matrices for layer-wise optimization. The required minimum number of nodes per layer is $2Q$. A potential initialization of $\mathbf{W}_l \in \mathbb{R}^{2Q \times 2Q}$ for $l=2,3,\hdots$  is $
\mathbf{W}_l = \mathbf{V}_m \mathbf{C} \mathbf{O}_{l-1}^{\star}$. We now show a limitation. While we have a lower limit on number of nodes for all the layers as $n_{min} = 2Q$, we lack an analytical setting for the upper limit $n_{max}$. The setting of $n_{max}$ remains as an experimental choice.

\subsection{Convex relaxation and structured weight matrix}

A practical system/algorithm establishes a trade-off between complexity and performance. Complexity includes modeling complexity and computational complexity. Modeling complexity refers to the structure of a system and the number of parameters in the system. Computational complexity refers to the computational requirement for generation of the structure and learning of the parameters. Henceforth we assume that $\mathbf{C}$ is an identity matrix to reduce complexity. We remove the use of $\mathbf{C}$ in LFP holding SLFN system. In that case, the parameter $m$ no longer plays any role and a feasible set of regularization parameters is
\begin{eqnarray}
\begin{array}{l}
n_{min} = 2Q, \\ 
\epsilon = 2Q, \,\, \mathrm{and} \\
\nu = 2Q\epsilon = (2Q)^2.
\end{array}
\end{eqnarray}

To decide the number of nodes in the $l$'th layer, we start with $n_l = 2Q$ and then increase $n_l$ until the cost \eqref{eq:LayerWise_CostFunction} saturates. It is straightforward to show that the increase in $n_l$ leads to the monotonically non-increasing cost for the SLFN. That means the optimized cost for $n_l =2Q+1$ is less than or equal to the optimized cost for $n_l = 2Q$. At the starting value $n_l = 2Q$, we set $\mathbf{W}_l = \mathbf{V}_Q \mathbf{O}_{l-1}^{\star}$ as initialization and solve \eqref{eq:LayerWise_Optimization_Problem} for every $n_l$ when $n_l$ increases. We stop when the cost minimization shows a saturation trend. 

Optimization of \eqref{eq:LayerWise_Optimization_Problem} for a chosen $n_l$ with respect to $\mathbf{O}_l$ and $\mathbf{W}_l$ is non-convex. We can use alternating optimization or gradient search. Use of alternating optimization or gradient search for the range of nodes $2Q \leq n_l \leq n_{max}$ is computationally intensive. Therefore we take two major practical steps for every $n_l$, discussed below.

\begin{enumerate}
	\item Convex relaxation: We construct $\mathbf{W}_l$ appropriately and fix it. The construction ensures that $n_l \geq n_{min} = 2Q$ and $\mathcal{C}_l \leq \mathcal{C}_{l-1}^{\star}$. We optimize $\mathbf{O}_l$ explicitly. This leads to a significant reduction in computational complexity. As we construct $\mathbf{W}_l$ and fix it, we remove the constraint $\| \mathbf{W}_l \|_F^2 \leq \nu$ in the optimization problem~\eqref{eq:LayerWise_Optimization_Problem}. Then, for the $l$'th layer, the optimization problem~\eqref{eq:LayerWise_Optimization_Problem} is relaxed to the following convex optimization problem:
	\begin{eqnarray}
	\underset{\boldsymbol{ \mathbf{O}_l}}{\arg\min} \,\, \mathcal{C}_l \,\,\,\, \mathrm{such \,\, that} \,\,
	\begin{array}{l}
	\| \mathbf{O}_l \|_F^2 \leq \epsilon'=\alpha 2Q, 
	\end{array}
	\label{eq:LayerWise_ConvexOptimization_Problem}
	\end{eqnarray}
	where $\mathcal{C}_l  = \frac{1}{J} \sum_{j=1}^J \| \mathbf{t}^{(j)} - \mathbf{O}_l \, \mathbf{y}_l^{(j)}  \|^2$, $\epsilon'=\alpha \epsilon$ and $\alpha \geq 1$ is a parameter that we set experimentally. The choice of $\alpha$ decides a size of feasible set. The vector $\mathbf{y}_l$ can be computed for a fixed $\mathbf{W}_l$.
	%We do not intend cross-validation to choose $\alpha$. For our experiments in Section~\ref{sec:Experimental_evaluations} we choose $\alpha = 2$ for many datasets. This is an arbitrary choice.  
	
	\item Use of a random matrix in construction of $\mathbf{W}_l$: Let us use $\mathbf{W}_{l,n_l}$ to denote the $\mathbf{W}_{l} \in \mathbb{R}^{n_l \times n_{l-1}}$ matrix to show dependency on $n_l$. Starting with $\mathbf{W}_{l,n_l=2Q} = \mathbf{V}_Q \mathbf{O}_{l-1}^{\star}$, we construct $\mathbf{W}_{l,n_l+1}$ matrix as follows
	\begin{eqnarray}
	\mathbf{W}_{l,n_l+1} = 
	\left[
	\begin{array}{c}
	\mathbf{W}_{l,n_l} \\
	\mathbf{r}
	\end{array}
	\right],
	\end{eqnarray}
	where $\mathbf{r}$ is a random instance based row vector. For the use of $\mathbf{W}_{l,n_l}$, let us use $\mathcal{C}_{l,n_l}^{\star}$ to denote the optimal cost achieved by solving~\eqref{eq:LayerWise_ConvexOptimization_Problem}. The above recursive construction of $\mathbf{W}_{l,n_l}$ guarantees monotonically non-increasing cost if we solve~\eqref{eq:LayerWise_ConvexOptimization_Problem} as the number of nodes increases. That means, we have
	\begin{eqnarray}
	\mathcal{C}_{l,n_l+1}^{\star} \leq \mathcal{C}_{l,n_l}^{\star} \leq \mathcal{C}_{l-1}^{\star}.
	\end{eqnarray}
	Similarly, we can increase the number of nodes by a step $\Delta$ and then construct 
	\begin{eqnarray}
	\mathbf{W}_{l,n_l+\Delta} = 
	\left[
	\begin{array}{c}
	\mathbf{W}_{l,n_l} \\
	\mathbf{R}
	\end{array}
	\right],
	\label{eq:StructureOfWeightMatrix_For_DeltaIncrease}
	\end{eqnarray}
	where $\mathbf{R} \in \mathbb{R}^{\Delta \times n_{l-1}}$ is a random instance based matrix. We have $\mathcal{C}_{l,n_l+\Delta}^{\star} \leq \mathcal{C}_{l,n_l+1}^{\star} \leq \mathcal{C}_{l,n_l}^{\star} \leq \mathcal{C}_{l-1}^{\star}$. We can draw components of $\mathbf{r}$ or $\mathbf{R}$ from iid Gaussian distribution or uniform distribution. 
\end{enumerate}
\begin{myremark}
	The structure of $\mathbf{W}_{l,n_l}$ matrix for $2Q \leq n_l \leq n_{max}$ follows a recursive relation as follows
	\begin{eqnarray}
	\mathbf{W}_{l,n_l} = 
	\left[
	\begin{array}{c}
	\mathbf{W}_{l,n_l-1} \\
	\mathbf{r}
	\end{array}
	\right] =
	\left[
	\begin{array}{c}
	\mathbf{V}_Q \mathbf{O}_{l-1}^{\star} \\
	\mathbf{R}'
	\end{array}
	\right] 
	\label{eq:StructureOfWeightMatrix}
	\end{eqnarray} 
	where $\mathbf{R}' \in \mathbb{R}^{(n_l - 2Q) \times n_{l-1}}$ is an instance of random matrix. 
\end{myremark}

\begin{figure*}[t!]
	\centering
	\def\svgwidth{\linewidth}
	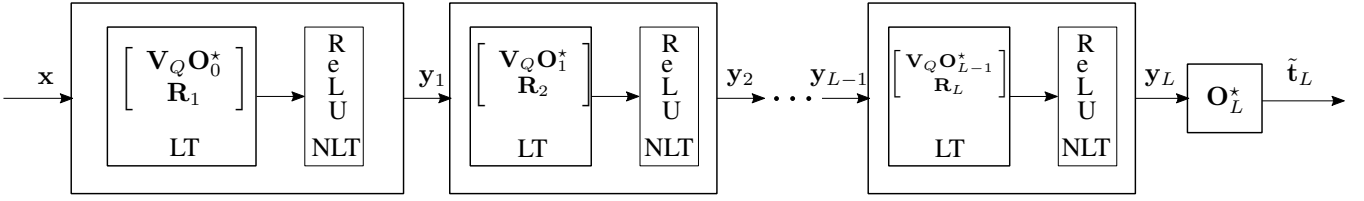	
	\caption{The architecture of a multi-layer SSFN with $L$ layers and its signal flow diagram. LT stands for \emph{linear transform} (weight matrix) and NLT stands for \emph{non-linear transform} (activation function). We use ReLU activation function.}
	\label{fig:MultiLayerPLN}
	%	\vspace{-12pt}
\end{figure*}

\begin{myremark}
The matrix $\mathbf{R}'$ in \eqref{eq:StructureOfWeightMatrix} can have a covolutional structure. In that case, we first choose a random instance of row vector and then form $\mathbf{R}'$ matrix as a circulant matrix from the row vector. Further, in lieu of random matrix instance $\mathbf{R}'$, it is possible to use row vectors from popular fixed transforms, such as discrete cosine transform (DCT), Wavelets, etc. The row vectors also can be impulse response of filters derived from filter banks, such as time-frequency analysis motivated filter banks, Gabor filters, visually and auditory response motivated filter banks, etc.
\end{myremark}
The prospect of fixed tranforms or convolutional structure or impulse response of filter banks to construct the $\mathbf{R}'$ part in an weight matrix is not investigated in this article. We continue with the use of random instance, shown in \eqref{eq:StructureOfWeightMatrix}. Using appropriate notations in \eqref{eq:StructureOfWeightMatrix}, the weight matrix for the $l$'th layer is written as
\begin{eqnarray}
	\mathbf{W}_{l} = 
	\left[
	\begin{array}{c}
	\mathbf{V}_Q \mathbf{O}_{l-1}^{\star} \\
	\mathbf{R}_l
	\end{array}
	\right]  \in \mathbb{R}^{n_l \times n_{l-1}},
	\label{eq:StructureOfWeightMatrix_LthLayer}
	\end{eqnarray} 	
	where $\mathbf{R}_l \in \mathbb{R}^{(n_l - 2Q) \times n_{l-1}}$ is the instance of random matrix. 
%We now discuss on a rank deficient aspect of weight matrices.
%
%\begin{myprop}
%	The weight matrix $\mathbf{W}_{l} \in \mathbb{R}^{n_l \times n_{l-1}}$ has the structure shown in \eqref{eq:StructureOfWeightMatrix}. The weight matrix is rank-deficient as follows
%	\begin{eqnarray}
%	\mathrm{rank}(\mathbf{W}_l) \leq \mathrm{min}(n_l , n_{l-1}) - Q.
%	\end{eqnarray}
%\end{myprop}
%
%\noindent Proof: $\mathrm{rank}(\mathbf{V}_Q \mathbf{O}_{l-1}^{\star}) \leq \min(\mathrm{rank}(\mathbf{V}_Q),\mathrm{rank}(\mathbf{O}_{l-1}^{\star})) \leq Q$. Let us first consider the case $n_l \leq n_{l-1}$. In that case, we deal with row rank  of $\mathbf{W}_l$ as $\mathrm{rank}(\mathbf{W}_l) \leq \mathrm{rank}(\mathbf{V}_Q \mathbf{O}_{l-1}^{\star}) + \mathrm{rank}(\mathbf{R}') \leq Q + (n_l - 2Q) = n_l - Q$. Now, we consider the case $n_l \geq n_{l-1}$. In this case, we deal with column rank of $\mathbf{W}_l$. For any matrix, column rank is equal to row rank. We have $n_l \geq n_{l-1}$ and $\mathrm{row \, rank}(\mathbf{W}_l) \leq n_{l-1} - Q$. \hfill $\blacksquare$
The architecture of SSFN and its signal flow diagram are shown in Figure \ref{fig:MultiLayerPLN}. 

While we used a random matrix instance as a part of an weight matrix, it is possible to learn (re-optimize) the full weight matrices for all the existing layers when we add a new node or a layer. The learning can be done using a gradient search. That will lead to an optimized system till the latest addition takes place. For every latest addition, we can re-optimize all the existing weight matrices. We did not pursue this re-optimization for every new addition of a node or a set of $\Delta$ nodes or a layer, as this re-optimization approach requires a significant  computational resource.

% The figure can be compared with the general structure of a feed-forward neural network illustrated in Figure~\ref{fig:Multi_layer_ANN}. Note the use of `blue' color circles to denote estimated target vector for each layer.

\medmuskip=-2mu
\medmuskip=4mu

\subsection{Advantage of using random instance in weight matrix formation and sequential learning for optimization}
The number of parameters is increasing as we add layers and nodes in SSFN. Weight matrix for $l$'th layer has the size $n_l \times n_{l-1}$. The total number of (scalar) parameters in the weight matrices for an $L$-layer feed-forward neural network is $\sum_{l=1}^L (n_{l-1} n_l)$. For SSFN, the structure of an weight matrix is shown in \eqref{eq:StructureOfWeightMatrix} and we are learning a part of it. The optimized part is $\mathbf{O}_{l-1}^{\star}$ of size $Q \times n_{l-1}$. Therefore, we are learning $\sum_{l=1}^L (Qn_{l-1})=Q\sum_{l=1}^L n_{l-1}$ parameters in total. Assuming $Q \ll n_l$, we have a significantly lower number of parameters to learn compared to the total number of parameters $\sum_{l=1}^L (n_{l-1} n_l)$. We can hope that this aspect of `learning a low number of parameters' brings an inherent regularization effect in SSFN.

We now discuss the advantage of sequential learning for increase in size of SSFN against possible methods that do not follow a sequential learning approach. Example of a possible method can be as follows. We could have started with a large size network and prune nodes and layers. Let us first consider pruning of layers. We start with a deepest network comprising of $L_{max}$ layers. Assume that the deepest network is already optimized using back propagation. Then we remove the last layer of the deepest network, optimize the cost function for the pruned network and check improvement of the cost due to the reduction of model complexity. The improvement can be tested on a validation dataset. If the improvement is reasonable then we continue similarly to prune the current last layer. Next, we consider pruning of nodes. 
%It is possible that nodes across layers will vary due to pruning. 
Pruning of nodes can be realized using a sparsity penalty on rows of an weight matrix and then combine the penalty in the cost optimization, for example, use of $\ell_1$-norm based penalty. Another example for pruning can be based on computation of statistical variance of signals in nodes. Low variance nodes can be pruned to achieve an appropriate network size. Pruning also can be done for those nodes that do not lead to a significant change in the optimized cost. In this case nodes of a layer can be ordered according to their influence on the cost and then pruned.

The above mentioned methods in the previous paragraph that do not follow sequential learning, require high computation. We start with a large size network that is already optimized. Optimization of a large size network is computationally demanding. Further, re-optimization in each step of pruning is also computationally demanding, and may be practically a daunting task. Our sequential learning approach is computationally simple.

\subsection{Low complexity convex optimization}

For the $l$'th layer, we need to solve the optimization problem \eqref{eq:LayerWise_ConvexOptimization_Problem}. While the optimization problem is convex, a practical problem is computational complexity for a large amount of training data and high-dimensional feature vector, that means if $J$ and $n_l$ are large. Therefore, computationally simple solutions are in need. The optimization problem can be solved in two ways. For the first case, the constrained form \eqref{eq:LayerWise_ConvexOptimization_Problem} can be handled using a computationally simple convex optimization method called alternating-direction-method-of-multipliers (ADMM) \cite{BoydPCPE11}. For the second case, an unconstrained Lagrangian form can be handled using a regularized least-squares (Tikonov regularization). 

We first discuss the second case where we handle the unconstrained Lagrangian form of \eqref{eq:LayerWise_ConvexOptimization_Problem}, shown below
\begin{eqnarray}
\underset{\boldsymbol{ \mathbf{O}_l}}{\arg\min} \,\, \left\{ \frac{1}{J} \sum_{j=1}^J \| \mathbf{t}^{(j)} - \mathbf{O}_l \, \mathbf{y}_l^{(j)}  \|^2 + \lambda_l \| \mathbf{O}_l \|_F^2   \right\}.
\label{eq:TikonovRegularizationForm_for_Layers}
\end{eqnarray}
Here $\lambda_l$ is a regularization parameter. The above Tikonov regularization has a closed form solution. The parameter $\epsilon'$ in the optimization problem~\eqref{eq:LayerWise_ConvexOptimization_Problem} and the parameter $\lambda_l$ have an intrinsic relation. If $\epsilon'$ increases then $\lambda_l$ typically decreases. While we have apriori knowledge to set $\epsilon' = \alpha 2Q$, we do not know how to set the value of $\lambda_l$. A typical approach for the choice of $\lambda_l$ is cross-validation. Cross validation is computationally intensive. Instead a simple approach can be as follows. We start with a small $\lambda_l$ and solve the Tikonov regularization problem for a fixed increment of $\lambda_l$ as a grid search. We stop the grid search when we see a saturation in decreasing trend of cost with the constraint $\| \mathbf{O}_l \|_F^2 \leq \epsilon'=\alpha 2Q$. While grid search is a simple approach, the problem is that it requires to solve Tikonov regularization several times. This might be a problem for a large amount of training data.

Alternatively we address the optimization problem \eqref{eq:LayerWise_ConvexOptimization_Problem} directly using ADMM.
ADMM is an iterative algorithm, more familiar in the parlance of distributed convex optimization  \cite{BoydPCPE11}. Apart from computational complexity, the use of ADMM can handle a distributed solution, for example, if the full training dataset is not in a single place, but distributed in several processing units. This leads to easy parallelism across multiple processors in computers. 
%Dropping the subscript $l$, and defining new matrices $ \mathbf{T}=[\mathbf{t}^{(1)}, \mathbf{t}^{(2)},\cdots,\mathbf{t}^{(J)}] $ and $ \mathbf{Y}=[\mathbf{y}^{(1)}, \mathbf{y}^{(2)},\cdots,\mathbf{y}^{(J)}] $, we can rewrite the optimization problem \eqref{eq:OptimizationProblem_PLN} for $p=2$ in the following constrained least-squares problem
%\begin{equation}
%\label{eq:ls}
%\min_{\Ob}~\frac{1}{2}\nf{\Tb-\Ob\Yb}^2~~~\mbox{s.t.}~~~\|\Ob\|_q \le \epsilon_o,
%\end{equation}
%where $\epsilon_o \triangleq \alpha^{\frac{1}{q}} \| \mathbf{U}_{Q} \|_q$ and $\|.\|_F$ denotes Frobenius norm.
To use ADMM, let us define new matrices $ \mathbf{T}=[\mathbf{t}^{(1)}, \mathbf{t}^{(2)},\cdots,\mathbf{t}^{(J)}] $ and $ \mathbf{Y}_l=[\mathbf{y}_l^{(1)}, \mathbf{y}_l^{(2)},\cdots,\mathbf{y}_l^{(J)}]$. We rewrite the optimization problem \eqref{eq:LayerWise_ConvexOptimization_Problem} in the following constrained form
\begin{equation}
\label{eq:ls}
\min_{\mathbf{O}_l}~\|\mathbf{T}-\mathbf{O}_l\mathbf{Y}_l\|_F^2~~~\mbox{such that}~~~\|\mathbf{O}_l\|_F \le \epsilon_{\alpha},
\end{equation}
where $\epsilon_{\alpha} \triangleq (\alpha2Q)^{\frac{1}{2}}$. 
To solve the above problem using ADMM, we consider the following equivalent form of \eqref{eq:ls}:
\begin{equation}
\label{eq:ls1}
\min_{\mathbf{O}, \mathbf{Q}}~\|\mathbf{T}-\mathbf{O}\mathbf{Y}\|_F^2~~~\mbox{s.t.}~~~\| \mathbf{Q} \|_F \leq \epsilon_{\alpha},~~\mathbf{Q}=\mathbf{O},
\end{equation}
where we drop the subscript $l$ for notational clarity.
Then, the ADMM iterations for solving the optimization problem would be as follows
\begin{equation} 
\label{eq:ls2}
\begin{cases}
\mathbf{O}(k \! + \! 1) \! = \! \underset{\mathbf{O}}{\arg\min} \|\mathbf{T} \! - \! \mathbf{O}\mathbf{Y}\|_F^2 \!+\! \frac{1}{\mu}\|\mathbf{Q}{(k)} \! - \!\mathbf{O} \!+\! \mathbf{\Lambda}{(k)}\|_F^2\\
\mathbf{Q}{(k \! + \! 1)} \! = \! \underset{\mathbf{Q}}{\arg\min} \|\mathbf{Q} \!-\! \mathbf{O}{(k \! + \! 1)} \!+\! \mathbf{\Lambda}{(k)}\|_F^2~\mbox{s.t.}~ \! \|{\mathbf{Q}}\|_F \! \le \! \epsilon_{\alpha} \\
\mathbf{\Lambda}{(k \! + \! 1)}  =\! \mathbf{\Lambda}{(k)}+\mathbf{Q}{(k \! + \! 1)}-\mathbf{O}{(k \! + \! 1)},
\end{cases}
\end{equation}
where $k$ denotes iteration index of ADMM, $ \mu>0 $ controls convergence rate of ADMM, and $\mathbf{\Lambda}$ stands for a Lagrange multiplier matrix. Noting that the two subproblems in \eqref{eq:ls2} have closed-form solutions, the ADMM steps are
\begin{equation}
\label{eq:ls3}
\begin{cases}
\mathbf{O}{(k+1)}=\big(\mathbf{T}\mathbf{Y}^T+\frac{1}{\mu}(\mathbf{Q}{(k)}+\mathbf{\Lambda}{(k)}\big)\cdot(\mathbf{Y}\mathbf{Y}^T+\frac{1}{\mu}\mathbf{I})^{-1}\\
\mathbf{Q}{(k+1)}=\mathcal{P}_{\mathcal{C}_q}(\mathbf{O}{(k+1)}-\mathbf{\Lambda}{(k)})\\
\mathbf{\Lambda}{(k+1)}=\mathbf{\Lambda}{(k)}+\mathbf{Q}{(k+1)}-\mathbf{O}{(k+1)}, 
\end{cases}
\end{equation}
in which, $ \mathcal{C}_q\triangleq \{ \mathbf{Q}\in\mathbb{R}^{Q\times n}:~ \| \mathbf{Q} \|_F \le \epsilon_{\alpha}\} $, and $ \mathcal{P}_{\mathcal{C}_q} $ performs projection onto $ \mathcal{C}_q $. The projection in \eqref{eq:ls3} has a closed-form solution, shown below
\begin{equation}
\mathcal{P}_{\mathcal{C}_q}(\mathbf{Q})=\left\{
\begin{array}{ll}
\mathbf{Q}\cdot(\frac{\epsilon_{\alpha}}{\|\mathbf{Q}\|_F}) & : \|\mathbf{Q}\|_F> \epsilon_{\alpha}\\
\mathbf{Q} & : \mbox{otherwise}.
\end{array}
\right.
\end{equation} As initial conditions for iterations, we set $\mathbf{Q}_{0}$ and $\mathbf{\Lambda}_{0}$ as zero matrices. The parameters to choose are $\mu$ and an upper limit on iterations denoted by $k_{max}$. The choice of $\mu$ has a high influence on the convergence rate of ADMM and the final solution. The parameter $\mu$ is chosen by hand-tuning. Note that the matrix inversion in \eqref{eq:ls3} is independent of the iterations, and as such it can be precomputed to save computations. In case of a training data limited scenario, when $ \mathbf{Y} $ is a tall matrix, we can invoke the Woodbury matrix identity to take the inverse of $ \mathbf{Y}^T\mathbf{Y} $ instead of $ \mathbf{Y}\mathbf{Y}^T $. 
%\begin{myremark}
%The matrix inversion in \eqref{eq:ls3} is independent of the iterations, and as such, it can be precomputed to save computations. Furthermore, when $ \Yb $ is a tall matrix, we can invoke the Woodbury matrix identity to take the inverse of $ \Yb^T\Yb $ instead of $ \Yb\Yb^T $. 
%\end{myremark}
%\begin{myremark}
%For $ q=1 $ ($ \ell_1 $ norm) and $ q=2 $ (Frobenius norm), the projection in \eqref{eq:ls3} has closed-form solutions. For $ q=2 $ we have
%\begin{equation}
%\Pc_{\Cc_q}(\Qb)=\left\{
%\begin{array}{lr}
%\Qb\cdot(\frac{\epsilon_o}{\nf{\Qb}}) & : \nf{\Qb}> \epsilon_o\\
%\Qb & : \mbox{otherwise}.
%\end{array}
%\right.
%\end{equation}
%\end{myremark}
%\begin{myremark}
%The projection in \eqref{eq:ls3} has a closed-form solution, shown below
%\begin{equation}
%\Pc_{\Cc_q}(\Qb)=\left\{
%\begin{array}{ll}
%\Qb\cdot(\frac{\epsilon_o}{\nf{\Qb}}) & : \nf{\Qb}> \epsilon_o\\
%\Qb & : \mbox{otherwise}.
%\end{array}
%\right.
%\end{equation}
%\end{myremark}

\subsection{SSFN learning algorithm and further optimization}
\label{subsec:PLN algorithm_and_further_optimization}

In the SSFN construction, we add layers one-by-one and nodes per layer in a step-wise manner. Construction of SSFN is shown in algorithm~\ref{alg:PLN_algorithm}. 
In the algorithm, for the first layer $l=1$, we have the parameter $\mathbf{O}_0^{\star}$ to construct the weight matrix $\mathbf{W}_1$. How to have an appropriate $\mathbf{O}_0^{\star}$? For the $0$'th layer we have $\mathbf{y}_0 = \mathbf{x}$, and we set the parameter $\mathbf{O}_0^{\star}$ using a regularized least-squares (Tikonov regularization), as follows 
\begin{eqnarray}
\underset{\boldsymbol{ \mathbf{O}_0}}{\arg\min} \,\, \frac{1}{J} \sum_{j=1}^J \| \mathbf{t}^{(j)} - \mathbf{O}_0 \, \mathbf{y}_0^{(j)}  \|^2 + \lambda_0 \| \mathbf{O}_0 \|_F^2.  
\label{eq:TikonovRegularizationForm_for_TheFirstLayer} 
\end{eqnarray}
We use cross-validation to set $\lambda_0$. 
Therefore, SSFN is expected to perform better than the regularized least-squares. The $\{ \mathbf{O}_l^{\star} \}_{l=1}^L$ matrices are learned by solving \eqref{eq:LayerWise_ConvexOptimization_Problem} using ADMM. ADMM has two parameters $\mu$ and $k_{max}$ to set.

In the SSFN, we set $L_{max}$, $n_{max}$ and $\alpha$. Then we set $\Delta$ as the number of nodes that we increase at a step for each layer. We use two more parameters $\eta_{node}$ and $\eta_{layer}$ for the stopping criteria. We stop increase in node for the $l$'th later if $\frac{C_{n_l}^{\star} - C_{n_l-\Delta}^{\star}}{C_{n_l-\Delta}^{\star}} < \eta_{node}$, that means when the cost shows a saturation trend. Similarly we stop increase in layer if $\frac{C_{l}^{\star} - C_{l-1}^{\star}}{C_{l-1}^{\star}} < \eta_{layer}$. There is a practical step in the SSFN algorithm. For every layer, we normalize the subvector $\mathbf{R}' \mathbf{y}_{l-1}$ to unit $\ell_2$-norm as $\frac{\mathbf{R}' \mathbf{y}_{l-1}}{\| \mathbf{R}' \mathbf{y}_{l-1} \|}$. This normalization step helps to arrest energy increase of signal flow through the successive layers of SSFN.

Once the process of increase in size of SSFN is over, we have a network structure of SSFN and its size. The SSFN has weight matrices $\{ \mathbf{W}_l \}_{l=1}^L$ and the output matrix $\mathbf{O}_L$ at the $L$'th layer. We then can re-optimize the weight matrices and the output matrix of SSFN using a gradient search for further optimization. We used a backpropagation algorithm from TensorFlow for optimization and learning of the parameters. The optimizer we use for backpropagation is called ADAM  \cite{ADAM_2015}. In ADAM, the learning rate of gradient search is found using a combination of hand tuning and cross-validation. Backpropagation is computationally complex. We call this backpropagation optimized SSFN as bSSFN. This bSSFN is expected to perform better than SSFN.

\begin{algorithm}[t]
	\caption{: Algorithm for construction of SSFN}\label{alg:PLN_algorithm}
	\mbox{Input: }
	\begin{algorithmic}[1]
		\STATE Training dataset $\mathcal{D} = \{ (\mathbf{x}^{(j)},\mathbf{t}^{(j)} ) \}_{j=1}^J$
		\STATE Parameters to set:
		\begin{enumerate}
			\item[(a)] $L_{max}$ \hfill (Maximum number of layers) 
			\item[(b)] $n_{max} \geq 2Q$ \hfill (Maximum number of nodes in a layer) 
			\item[(c)] $\alpha \geq 1$    \hfill (Parameter in \eqref{eq:LayerWise_ConvexOptimization_Problem})
			\item[(d)] $\Delta$ \hfill (Numbers of nodes to increase in a step)
			\item[(e)] $\mu$ and $k_{max}$ \hfill (Parameters in ADMM)
			\item[(f)] $\eta_{node}$ and $\eta_{layer}$ \hfill (Stopping thresholds)
		\end{enumerate}
	\end{algorithmic}
	\mbox{Regularized least-squares:}
	\begin{algorithmic}[1]
		\STATE $\mathbf{y}_0^{(j)} = \mathbf{x}^{(j)}$
		\STATE Solve \eqref{eq:TikonovRegularizationForm_for_TheFirstLayer} to find $\mathbf{O}_0^{\star}$ \hfill (Cross-validation for $\lambda_0$)
	\end{algorithmic}
	\mbox{Initialization:}
	\begin{algorithmic}[1]
		\STATE $l=0$  \hfill (Index for $l$'th layer)
	\end{algorithmic}
	\mbox{Estimating number of nodes and layers:}
	\vspace{-0.3cm}
	\begin{algorithmic}[1]
		\REPEAT 
		\STATE $l \leftarrow l+1$  \hfill (Increase in layers)
		\STATE $n_l = 2Q$ \hfill (Minimum number of nodes for all layers)
		\REPEAT
		\STATE $n_l \leftarrow n_l + \Delta$ \hfill (Increase in nodes)
		\STATE Construct $\mathbf{W}_{l,n_l}$ according to \eqref{eq:StructureOfWeightMatrix_For_DeltaIncrease} and \eqref{eq:StructureOfWeightMatrix}
		\STATE Find feature $\mathbf{y}_{l,n_l}^{(j)}$ \hfill (For $n_l$ nodes)
		\STATE Solve \eqref{eq:LayerWise_ConvexOptimization_Problem} to find $\mathbf{O}_{l,n_l}^{\star}$ \hfill (using ADMM)
		\UNTIL $\frac{C_{n_l}^{\star} - C_{n_l-\Delta}^{\star}}{C_{n_l-\Delta}^{\star}} < \eta_{node}$ and $n_l > n_{max}$
		\STATE $\mathbf{O}_{l}^{\star} \leftarrow \mathbf{O}_{l,n_l}^{\star}$, $C_{l}^{\star} \leftarrow C_{n_l}^{\star}$
		\UNTIL $\frac{C_{l}^{\star} - C_{l-1}^{\star}}{C_{l-1}^{\star}} < \eta_{layer}$ and $l > L_{max}$
	\end{algorithmic}
	\mbox{Output:}
	\begin{algorithmic}[1]
		\STATE Number of layers $L = l$ and number of nodes $\{ n_l \}_{l=1}^L$
		\STATE Weight matrices $\{ \mathbf{W}_{l} \}_{l=1}^L$
	\end{algorithmic}
\end{algorithm}

\section{Experimental Evaluations}
\label{sec:Experimental_evaluations}

Our experiments will consider: self size-estimation, low computational complexity requirement, limited human effort in tuning parameters, consistent performance in classification accuracy and estimated size, comparison with state-of-the-art, and finally a failure case with a mitigation approach.

\begin{table}[t]
	\centering
	\caption{Dataset for multi-class classification}
	\vspace{-2mm}
	\label{table:Database_for_classification}
	\setlength{\tabcolsep}{1pt}
	\begin{tabular}{ |c|c|c|c|c|c| } 
		\hline
		Dataset & {\begin{tabular}{@{}c@{}} Number of  \\ train data\end{tabular}}  & {\begin{tabular}{@{}c@{}}Number of  \\ test data\end{tabular}} & {\begin{tabular}{@{}c@{}}Input  \\ dimension ($\mathit{P}$)\end{tabular}}  & {\begin{tabular}{@{}c@{}} Number of  \\ classes ($\mathit{Q}$)\end{tabular}} & {\begin{tabular}{@{}c@{}}Random  \\ Partition\end{tabular}}\\
		\hline \hline 
		Vowel & 528 & 462 & 10 & 11 & No\\ 
		\hline
		%		Extended YaleB & 1600 & 800 & 504 & 38 & Yes\\ 
		%		\hline
		%		AR & 1800 & 800 & 540 & 100 & Yes\\ 
		%		\hline
		Satimage & 4435 & 2000 & 36 & 6 & No\\ 
		\hline
		%		Scene15 & 3000 & 1400 & 3000 & 15 & Yes\\ 
		%		\hline
		Caltech101 & 6000 & 3000 & 3000 & 102 & Yes\\ 
		\hline
		Letter & 13333 & 6667 & 16 & 26 & Yes\\ 
		\hline
		NORB & 24300 & 24300 & 2048 & 5 & No\\ 
		\hline
		Shuttle & 43500 & 14500 & 9 & 7 & No\\ 
		\hline
		MNIST & 60000 & 10000 & 784 & 10 & No\\ 
		\hline
		CIFAR-10 & 50000 & 10000 & 3072 & 10 & No \\
		\hline
		%		CIFAR-100 & 50000 & 10000 & 3072 & 100 & No \\
		%		\hline
	\end{tabular}
	\vspace{-0.3cm}
\end{table}

%\begin{table}[t]
%	\centering
%	\caption{Databases for regression}
%	\label{table:Database_for_regression}
%	\setlength{\tabcolsep}{2.5pt}
%	\begin{tabular}{ |c|c|c|c|c|c| } 
%		\hline
%		Database & {\begin{tabular}{@{}c@{}}$\#$ of  \\ train data\end{tabular}}  & {\begin{tabular}{@{}c@{}}$\#$ of  \\ test data\end{tabular}} & {\begin{tabular}{@{}c@{}}Input  \\ dimension ($\mathit{P}$)\end{tabular}}  & {\begin{tabular}{@{}c@{}}Target  \\ dimension ($\mathit{Q}$)\end{tabular}} & {\begin{tabular}{@{}c@{}}Random  \\ Partition 
%		\end{tabular}}\\
%		\hline \hline
%		Pyrim & 49 & 25 & 27 & 1 & Yes\\ 
%		\hline
%		Bodyfat & 168 & 84 & 14 & 1 & Yes\\ 
%		\hline
%		Housing & 337 & 169 & 13 & 1 & Yes\\ 
%		\hline
%		Strike & 416 & 209 & 6 & 1 & Yes\\ 
%		\hline 
%		Balloon & 1334 & 667 & 2 & 1 & Yes\\ 
%		\hline
%		Space-ga & 2071 & 1036 & 6 & 1 & Yes\\ 
%		\hline
%		Abalone & 2784 & 1393 & 8 & 1 & Yes\\ 
%		\hline
%		Parkinsons & 4000 & 1875 & 20 & 2 & Yes\\
%		\hline
%	\end{tabular}
%\end{table}

%We perform classification experiments in this section.

\subsection{Experimental setups}

\subsubsection{Datasets}
Table \ref{table:Database_for_classification} shows eight datasets that we use for experimental evaluations. These datasets are chosen due to their diversity in signals, popularity in literature and level of complexity for tasks.
% These databases are extensively used in practice.
%the literarture \cite{elob07,telmmp16,jlcksvd13,zdksvd10,Zldfs14,aksvd06,fboosting99}. 
The `vowel' dataset is for vowel recognition task (a speech recognition application) and all other seven datasets are for image object classification task. We test both speech recognition and image classification due to task diversity. In the Table \ref{table:Database_for_classification}, we show number of training data samples, number of test data samples, input signal dimension ($P$), number of classes ($Q$), and a column identifier as `random partition'. For a few datasets, the input signal dimension $P$ is small, say for the vowel dataset and letter dataset. We choose such datasets to accommodate low resolution data/features. Note that the number of training samples varies significantly across the datasets. 
%For example, the vowel dataset has a very limited number of training samples. 
For six datasets, we have access to the predefined training and test datasets. Caltech101 and Letter datasets do not have predefined training and test datasets. For the Caltech101 and Letter datasets, we create training and test datasets using random sampling from the full dataset. We mark the identifier `random partition' as `yes' for these two datasets. In the case of Caltech101 dataset, we use 3000-dimensional feature vectors suggested in \cite{Label_Consistent_KSVD_2013} for the proposed methods and image signals directly for evaluating a competitor method. Caltech101 dataset has images with varying pixel size. All images are downsampled to $128 \times 128$ pixel size for the competitor method. As we focus on classification, the target vector $\mathbf{t}$ is a $Q$-dimensional categorical variable, and we decide the class that corresponds to the coordinate of the highest amplitude scalar component of the predicted target vector $\tilde{\mathbf{t}}$. 

\subsubsection{Software and hardware} 
We use Matlab and Python for programming. We use a laptop and two servers for the hardware support. The laptop is used for SSFN. A server is used for back propagation based optimization in bSSFN. The laptop uses 2.6 GHz processor and 16 GB RAM, and the server uses multi-processors and 256 GB RAM. We trained and tested convolutional neural networks (CNN) for comparison. CNN is trained using a GPU enabled server. We used the Keras CNN example for implementation\footnote{Available at: \url{https://keras.io/examples/cifar10_cnn/}}. For back propagation in bSSFN, we used ADAM  \cite{ADAM_2015} from TensorFlow. SSFN training time is in the order of ten minutes using the laptop. Back propagation in bSSFN requires hours in the server. CNN training also took hours in the GPU enabled server.

\subsubsection{Reproducible research}
%In spirit of reproducible research, all software codes are posted in the web https://sites.google.com/site/saikatchatt/ and also in the web www.ee.kth.se/reproducible/. The codes can be used to reproduce some experimental results reported in this article.
Matlab and Python codes are available in https://sites.google.com/site/saikatchatt/ and www.ee.kth.se/reproducible/. 
%The codes can be used to reproduce some experimental results reported in this article.

\subsection{Experimental results}

\subsubsection{On self size-estimation, low computation, limited human effort, consistent performance}

%\begin{table}[t!]
%	\centering
%	\caption{Classification accuracy of SSFN}
%	\vspace{-2mm}
%	\label{table:Classification_performance_without_Zscore_saikat}
%	\setlength{\tabcolsep}{1.5pt}
%	\renewcommand{\arraystretch}{1.4}
%	\begin{tabular}{|c|c|c|c|c|c|} 
%		\hline
%		\multirow{2}{*}{Dataset} & Regularized LS & \multicolumn{4}{|c|}{SSFN}   \\ \cline{2-6}
%		& Accuracy         & Accuracy & $\lambda_0$ & $\mu$ & Other parameters \\ \hline
%		Vowel &  28.1  & 60.2 $\pm$ 2.4 & $10^2$ & $10^3$ &  \multirow{8}{*}{{\begin{tabular}{@{}c@{}} $k_{max} =100$ \\ $\alpha=2$ \\ $\!n_{max}\!-\!2Q \!= \!1000 \!$ \\ $\eta_{node} = 0.005$ \\ $\eta_{layer} = 0.1$ \\ $L_{max} = 20$ \\ $\Delta=50$ \end{tabular}} } \\ \cline{1-5}
%		Satimage & 68.1 & 89.9 $\pm$ 0.5 & $10^6$ & $10^5$ &  \\ \cline{1-5}
%		Caltech101 & 66.3 & 76.1 $\pm$ 0.8 & $5$ & $10^{-2}$ &  \\ \cline{1-5}
%		Letter & 55.0 & 95.7 $\pm$ 0.2 & $10^{-5}$ & $10^4$ &  \\ \cline{1-5}
%		NORB & 80.4 & 86.1 $\pm$ 0.2 & $10^2$ & $10^2$ &  \\ \cline{1-5}
%		Shuttle & 89.2 & 99.8 $\pm$ 0.1 & $10^5$ & $10^4$ &  \\ \cline{1-5}
%		MNIST & 85.3 & 95.7 $\pm$ 0.1 & $1$ & $10^5$ &  \\ \cline{1-5}
%		CIFAR-10 & 40.3 & 47.3 $\pm$ 0.2 & $10^8$ & $10^3$ &  \\  \hline
%		%		CIFAR-100 & 14.9 & 23.0 $\pm$ 0.2 & $10^8$ & 1 &  \\ \hline
%	\end{tabular}
%\end{table}

\begin{table*}[t!]
	\centering
	\caption{Classification accuracy of SSFN across 50 Monte-Carlo simulations, its complexity requirement (execution time in the laptop) and human effort required to set its parameters}
	\vspace{-2mm}
	\label{table:Classification_performance_without_Zscore_saikat}
	\setlength{\tabcolsep}{7.5pt}
	\renewcommand{\arraystretch}{1.4}
	\begin{tabular}{|c|c|c|c|c|c|c|} 
		\hline
		\multirow{2}{*}{Dataset} & Regularized LS & \multicolumn{5}{|c|}{SSFN}   \\ \cline{2-7}
		& Accuracy (in $\%$)   & Accuracy (in $\%$) & Complexity &\multicolumn{3}{|c|}{Parameters to set}  \\ \cline{5-7}
		& & (avg. $\pm$ std. dev.) & (average learning time & \multicolumn{2}{|c|}{Manual effort} & Limited manual effort for all other parameters  \\ \cline{5-6}
		& & &  in seconds) & $\lambda_0$ & $\mu$ & (chosen same across all the eight datasets)  \\ \hline
		Vowel &  28.1  & 60.2 $\pm$ 2.4 & 6 s & $10^2$ & $10^3$ &  \multirow{8}{*}{{\begin{tabular}{@{}c@{}} $k_{max} =100$ \\ $\alpha=2$ \\ $\!n_{max}\!-\!2Q \!= \!1000 \!$ \\ $\eta_{node} = 0.005$ \\ $\eta_{layer} = 0.1$ \\ $L_{max} = 20$ \\ $\Delta=50$ \end{tabular}} }  \\ \cline{1-6}
		Satimage & 68.1 & 89.9 $\pm$ 0.5 & 11 s & $10^6$ & $10^5$ &   \\ \cline{1-6}
		Caltech101 & 66.3 & 76.1 $\pm$ 0.8 & 84 s & $5$ & $10^{-2}$ &   \\ \cline{1-6}
		Letter & 55.0 & 95.7 $\pm$ 0.2 & 248 s &$10^{-5}$ & $10^4$ &  \\ \cline{1-6}
		NORB & 80.4 & 86.1 $\pm$ 0.2 & 145 s & $10^2$ & $10^2$ &   \\ \cline{1-6}
		Shuttle & 89.2 & 99.8 $\pm$ 0.1 & 61 s & $10^5$ & $10^4$ &   \\ \cline{1-6}
		MNIST & 85.3 & 95.7 $\pm$ 0.1 & 227 s & $1$ & $10^5$ &  \\ \cline{1-6}
		CIFAR-10 & 40.3 & 47.3 $\pm$ 0.2 & 206 s & $10^8$ & $10^3$ &   \\  \hline
		%		CIFAR-100 & 14.9 & 23.0 $\pm$ 0.2 & & $10^8$ & 1 &  \\ \hline
	\end{tabular}
\end{table*}

%\begin{table}[t!]
%	\centering
%	\caption{Classification accuracy of ELM and PLN. The parameters of PLN are as follows: $L_{max}=20$ , $\Delta=50$ , $k_{max} =100$ , $\alpha=2$ , $\!n_{max}\!-\!2Q \!= \!1000 \!$ , $\eta_{node} = 0.005$ , $\eta_{layer} = 0.1$. The number of nodes of ELM is $2 Q + 1000$.}
%	\label{table:Classification_performance_without_Zscore_saikat}
%	\setlength{\tabcolsep}{2.9pt}
%	\renewcommand{\arraystretch}{1.4}
%	\begin{tabular}{|c|c|c|c|c|c|c|} 
%		\hline
%		\multirow{2}{*}{Dataset} & Regularized LS & \multicolumn{3}{|c|}{PLN} & \multicolumn{2}{c|}{ELM} \\ \cline{2-7}
%		& Accuracy & Accuracy & $\lambda_0$ & $\mu$ & Accuracy & $\lambda$ \\ \hline
%		Vowel &  28.1  & 60.2 $\pm$ 2.4 & $10^2$ & $10^3$ & 52.8$\pm$1.6 & $10^{2}$ \\ \cline{1-7}
%		Satimage & 68.1 & 89.9 $\pm$ 0.5 & $10^6$ & $10^5$ & 83.9$\pm$0.4 & $10^2$ \\ \cline{1-7}
%		Caltech101 & 66.3 & 76.1 $\pm$ 0.8 & $5$ & $10^{-2}$ & 51.5$\pm$0.8 & $10$ \\ \cline{1-7}
%		Letter & 55.0 & 95.7 $\pm$ 0.2 & $10^{-5}$ & $10^4$ & 89.2$\pm$0.3 & $10$\\ \cline{1-7}
%		NORB & 80.4 & 86.1 $\pm$ 0.2 & $10^2$ & $10^2$ & 85.9$\pm$0.6 & $10^3$\\ \cline{1-7}
%		Shuttle & 89.2 & 99.8 $\pm$ 0.1 & $10^5$ & $10^4$ & 99.5$\pm$0.1 & $10^2$ \\ \cline{1-7}
%		MNIST & 85.3 & 95.7 $\pm$ 0.1 & $1$ & $10^5$ & 94.4$\pm$0.2 & $10^{-1}$ \\ \cline{1-7}
%		CIFAR-10 & 40.3 & 47.3 $\pm$ 0.2 & $10^8$ & $10^3$ & 44.0$\pm$0.4 & $1$ \\ \hline
%		%		CIFAR-100 & 14.9 & 23.0 $\pm$ 0.2 & $10^8$ & 1 &  \\ \hline
%	\end{tabular}
%\end{table}

\begin{table}[t!]
	\centering
	\caption{Size of SSFN for four randomly chosen Monte Carlo simulations among 50 simulations and four randomly chosen datasets to show consistent estimation of size}
	\vspace{-2mm}
	\label{table:SizeOfPLN}
	\setlength{\tabcolsep}{3pt}
	\renewcommand{\arraystretch}{1.2}
	\begin{tabular}{ |c|c|c|} 
		\hline
		Dataset & Arrangement of nodes across layers  & Accuracy\\
		\hline \hline 
		\multirow{4}{*}{Vowel} & 272-222-222-372-322-372-372-522-1022-922-72 & 59.50 \\
		& 272-172-322-272-322-322-472-572-722-1022-72 & 62.34 \\ 
		& 322-222-272-272-322-322-372-422-522-1022-72 & 62.55 \\ 
		& 272-222-222-272-322-322-372-422-522-1022-72 & 59.74 \\ 
		%& 272-272-222-322-372-372-322-372-422-1022-72 & 62.12 & \\ 
		\hline
		%		\multirow{5}{*}{Extended YaleB} & 676-876-1076-1076-126-126 & 98.13 \\
		%		& 576-926-1076-1076-126-126 & 97.13 \\ 
		%		& 626-876-1076-1076-126-126 & 98.75 \\ 
		%		& 676-926-1076-1076-126-126 & 97.50 \\ 
		%		& 626-876-1076-1076-126-126 & 97.75 \\ 
		%		\hline
		%		\multirow{5}{*}{AR} & 550-950-1200-1200-250-250 & 98.25 \\
		%		& 600-1050-1200-1200-250-250 & 98.00 \\ 
		%		& 550-1000-1200-1200-250-250 & 98.50 \\ 
		%		& 550-900-1200-1200-250-250 & 97.38 \\ 
		%		& 550-1100-1200-1200-250-250 & 98.88 \\ 
		%		\hline
		%		\multirow{5}{*}{Satimage} & 414-414-564-514-414 & 90.20 \\
		%		& 264-464-464-514-314 & 90.55 \\
		%		& 314-414-514-514-264 & 90.30 \\
		%		& 264-414-564-464-464 & 90.25 \\
		%		& 364-364-714-514-514-464 & 89.10 \\
		%		\hline
		%		\multirow{5}{*}{Scene15} & 1030-1030-1030-130-980-80 & 99.36 \\
		%		& 1030-1030-1030-1030-80-880-80 & 98.71 \\
		%		& 1030-1030-1030-1030-80-1030-80 & 99.07 \\
		%		& 1030-1030-1030-1030-80-930-80 & 99.43 \\
		%		& 1030-1030-1030-80-880-80 & 98.93 \\
		%		\hline
		\multirow{5}{*}{Caltech101} & 1204-604-454-404-504-454-454 & 73.83\\
		& 1204-604-404-404-504-454-454 & 73.31\\
		& 1204-654-454-454-454-454-404 & 73.67\\
		& 1204-604-454-404-454-454-454 & 73.26\\
		%		& 1204-604-404-454-504-454-404 & 73.09\\
		\hline
		\multirow{5}{*}{Letter} & 952-1052-1052-652-1052-302-1002-252 & 95.70 \\
		& 1052-1052-1052-702-1052-202 & 95.40\\
		& 952-1052-1052-852-652-1052-252 & 95.52\\
		& 1002-1052-1052-752-902-502-952-252 & 95.43\\
		%		& 952-1052-1052-352-1052-202 & 95.59 \\
		\hline
		%		\multirow{5}{*}{NORB} & 960-510-360-210 & 85.55 \\
		%		& 1010-510-510-210 & 85.35 \\
		%		& 1010-410 & 86.64 \\
		%		& 1010-410-410-160 & 85.63 \\
		%		& 1010-460 & 86.34 \\
		%		\hline
		%		\multirow{5}{*}{Shuttle} & 364-214-364-314 & 99.88 \\
		%		& 414-264-414-364-264 & 99.85 \\
		%		& 364-264-364-314 & 99.88 \\
		%		& 264-214-314-264 & 99.81 \\
		%		& 364-264-364-314 & 99.86 \\
		%		\hline
		\multirow{5}{*}{MNIST} & 1020-170-770-120 & 95.55\\
		& 1020-170-870-70 & 95.78\\
		& 1020-170-820-120 & 95.54\\
		& 1020-220-870-120 & 95.75\\
		%		& 1020-120-820-70 & 95.84\\
		\hline
	\end{tabular}
	%\vspace{-0.3cm}
\end{table}

%We begin with our first experiment to verify the hypothesis that SSFN can estimate its own size and the size of SSFN varies significantly across datasets. In this experiment, 
We begin with our first experiment where the classification accuracy of SSFN is reported in Table~\ref{table:Classification_performance_without_Zscore_saikat}. In the table, performance of the regularized least-squares (LS) is reported for baseline comparison. SSFN provides significant performance improvement than regularized LS. SSFN architecture starts with regularized LS in its first layer and then grows to a multi-layer structure. All the design parameters of SSFN are also shown in Table~\ref{table:Classification_performance_without_Zscore_saikat}. The parameter $\lambda_0$ is common to both the SSFN and the regularized LS. The choice of $\mu$ influences convergence of ADMM in SSFN. These two parameters - $\lambda_0$ and $\mu$ - are set using a combination of cross-validation and manual effort. Note that, for this experiment, all the other design parameters of SSFN are deliberately kept same for all the eight datasets. We did not tune them and this can be considered as a limited human effort. We use random matrix instances in SSFN and hence, show average performance over 50 Monte Carlo simulations. The standard deviation of accuracy for the Monte Carlo simulations is also reported in the table with the notation `$\pm$'. The standard deviation is low, signifying consistent classification performance. The table also shows computational resource required to design SSFN architecture and learning its parameters. This computational complexity is low compared to training many contemporary neural networks, for example CNN. 

\begin{figure*}[t!]
	\centering
	\includegraphics[width=1\textwidth, trim = 95 0 95 0,clip]{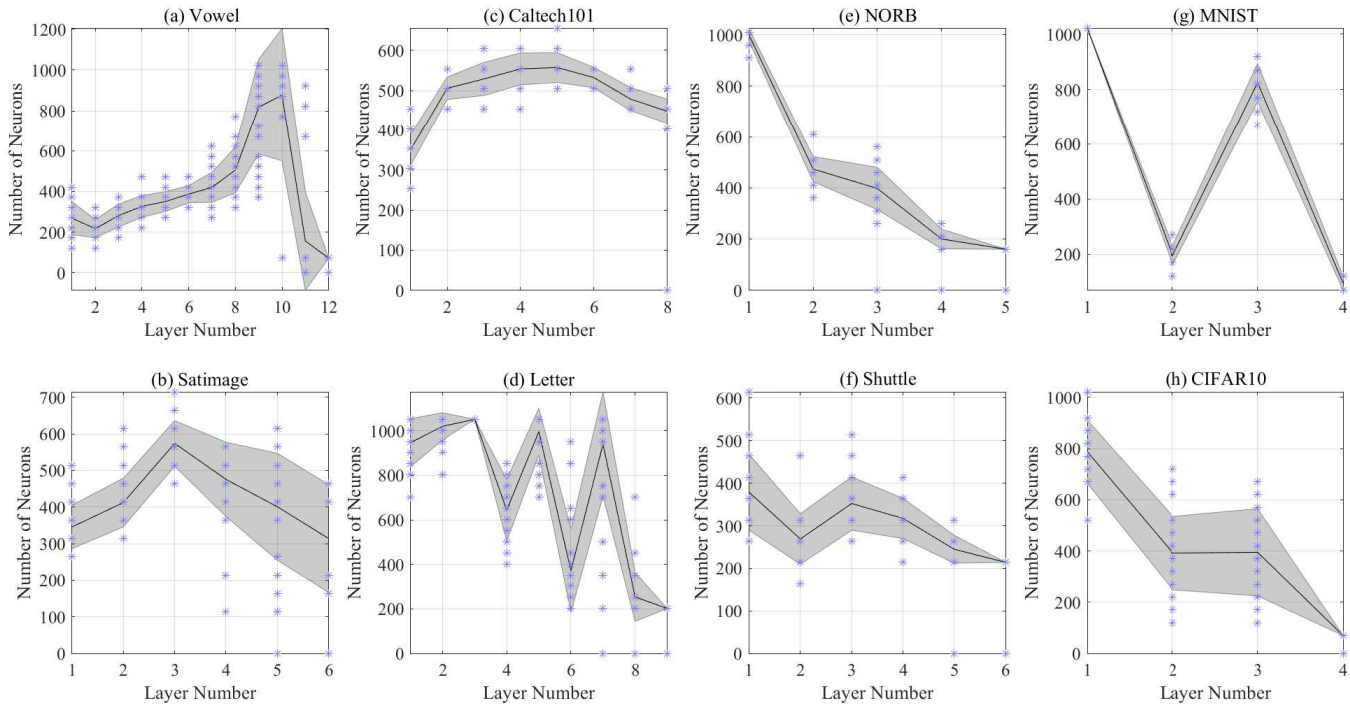}
	\vspace{-0.3cm}	
	\caption{Number of nodes (neurons) per layer versus the layer number in SSFN. We have 50 Monte Carlo simulations. A `blue star' mark denotes an instance of number of nodes at a layer. The instance corresponds to an SSFN instance due to the use of random matrix instances in weight matrices. The solid line and shaded region illustrate the mean and the standard deviation of the number of neurons per layer for Monte Carlo simulations, respectively. If the number of nodes is equal to zero, then, there is no layer for the corresponding SSFN instance. Using the Monte Carlo simulations, the average number of layers in SSFN for Vowel, Satimage, Caltech101, Letter, NORB, Shuttle, MNIST, and CIFAR10 datsets are found to be 10.9, 4.8, 7.3, 7.9, 3.4, 4.5, 4, and 3.6, respectively. We get fractional numbers, such as 10.9 layers, due to computing average number of layers across Monte Carlo simulations. The figure shows that the size of SSFN varies significantly across the eight datasets.}
	\label{fig:PLN_Trials_size}
\end{figure*}

We now discuss on size of SSFN and how SSFN estimates its own size. The size of SSFN for four datasets is shown in Table~\ref{table:SizeOfPLN}. %Parameters of PLN are according to the Table~\ref{table:Classification_performance_without_Zscore_saikat}. 
For each of the four datasets, we show the size for four Monte-Carlo simulations randomly chosen from the 50 Monte-Carlo simulations. Suppose we consider MNIST dataset. In Table~\ref{table:SizeOfPLN}, the entry `1020-170-770-120' means that the SSFN has four layers, and the number of nodes for the first, second, third and fourth layer is 1020, 170, 770 and 120, respectively. 
It is interesting to observe that the size of SSFN remains similar across Monte Carlo simulations for a dataset. This can be considered as consistent size estimation - the neural network system architecture does not vary randomly across Monte-Carlo simulations.
%For example, all the four instances of SSFN for MNIST dataset are similar to each other. 
Pictorial visualization of SSFN size for 50 Monte Carlo simulations is shown in Figure~\ref{fig:PLN_Trials_size}. It is interesting to observe how the size varies across datasets and across Monte Carlo simulations for a dataset. For example, let us consider the Vowel dataset where the number of nodes $n_l$ for the $l$'th layer slowly increases with the layer number $l$, and then suddenly decreases. For the Letter and MNIST datasets, the number of nodes $n_l$ shows an increase-and-decrease trend, almost alternatively with respect to the layer number $l$. An arrangement of high-low-high number of nodes in consecutive layers reminds us the use of an autoencoder architecture. Table~\ref{table:SizeOfPLN} and Figure~\ref{fig:PLN_Trials_size} show that SSFN can estimate its own size in a consistent manner for a dataset and the size varies significantly across datasets. 

\begin{figure*}[t!]
	\centering
	\includegraphics[width=1\textwidth, trim = 95 0 95 0,clip]{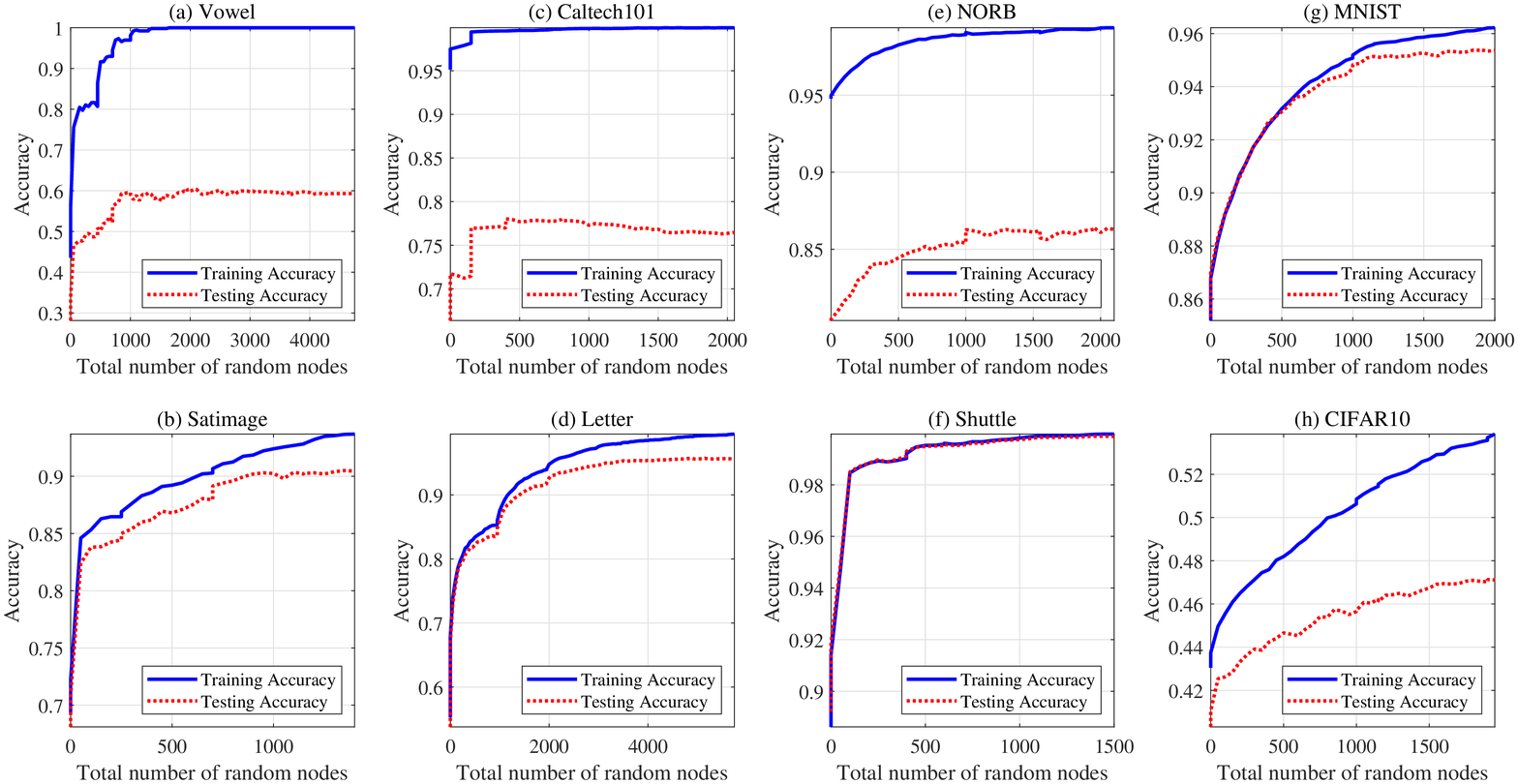}
	\vspace{-0.5cm}
	\caption{Training and testing accuracy against size of SSFN. The size of an $l$-layer SSFN is represented by the total number of random matrix instance based nodes, counted as $\sum_{l'=1}^l (n_{l'}-2Q)$. Plots are shown for all the eight datasets.}
	\label{fig:PerformanceImprovement_with_PLN_size}
	%\vspace{-0.3cm}
\end{figure*}

%\subsubsection{Improvement in classification accuracy}
We now discuss the improvement of classification accuracy with the increase in size of SSFN. 
%This is in line with the success of deep neural networks comprised of many layers. 
SSFN starts with the regularized LS and then grows its size with addition of nodes and layers. The accuracy improvement for all the eight datasets is shown in Figure~\ref{fig:PerformanceImprovement_with_PLN_size}. We plot accuracy versus number of nodes $\sum_{l'=1}^l (n_{l'}-2Q)$. The number of nodes $\sum_{l'=1}^l (n_{l'}-2Q)$ is associated with the random instance parts of weight matrices. This number of nodes represents the increase in size of SSFN. In the Figure~\ref{fig:PerformanceImprovement_with_PLN_size}, we show accuracy for training set and test set for each of the eight datasets. We observe that the training accuracy improves with increase in size of SSFN. 

In the the second experiment we see the effects of manual effort (hand tuning) for some of the parameters of SSFN. Hand tuning is an art of design. It is driven by intuition and trial-and-error. We tune the number of random instance based nodes $(n_{max}-2Q)$, the stopping parameter $\eta_{layer}$ and the step  $\Delta$ for increase in number of nodes. The other parameters remain same as in Table~\ref{table:Classification_performance_without_Zscore_saikat}. We show performance of hand-tuned SSFN (hSSFN) in Table~\ref{table:tuned_PLN_performance_classification_saikat} and observe that the hand tuning helps. For example, classification accuracy improves to $98\%$ for MNIST dataset.

\begin{table*}[t!]
	\centering
	\caption{Classification performance for SSFN and hand-tuned SSFN (hSSFN)}
	\vspace{-2mm}
	\setlength{\tabcolsep}{9pt}
	\renewcommand{\arraystretch}{1.3}
	\label{table:tuned_PLN_performance_classification_saikat}
	\begin{tabular}{|c|c|c|c|c|c|c|c|c|c|c|c|} 
		\hline
		\multirow{2}{*}{Dataset} & \multicolumn{1}{c}{Accuracy} & \multicolumn{1}{|c|}{Accuracy} & \multicolumn{9}{c|}{Parameters of hSSFN (some parameters are set using trial-and-error)} \\ \cline{4-12}
		& SSFN & hSSFN & $\lambda_{0}$ & $\mu$ & $k_{max}$ & $\alpha$ & $n_{max}-2Q$ & $\eta_{node}$ & $\eta_{layer}$ & $L_{max}$ & $\Delta$ \\
		\hline \hline
		Vowel & 60.2 $\pm$ 2.4 & 63.3 $\pm$ 1.5 & $10^{2}$ & $10^3$ & 100 & $2$ & $4000$ & $0.005$ & $0.05$ & $20$ & $500$ \\ 
		\cline{1-12}
		Satimage & 89.9 $\pm$ 0.5 & 90.8 $\pm$ 0.3 & $10^{6}$ & $10^5$ & 100 & $2$ & $4000$ & $0.005$ & $0.15$ & $20$ & $500$ \\ 
		\cline{1-12}
		Caltech101 & 76.1 $\pm$ 0.8 & 77.5 $\pm$ 0.7 & $5$ & $10^{-2}$ & 100 & $3$ & 20 & $0.005$ & $0.15$ & $20$ & $5$\\ 
		\cline{1-12}
		Letter & 95.7 $\pm$ 0.2 & 97.1 $\pm$ 0.3 & $10^{-5}$ & $10^4$ & 100 & $2$ & $4000$ & $0.005$ & $0.25$ & 20 & $500$  \\ 
		\cline{1-12}
		NORB & 86.1 $\pm$ 0.2 & 87.8 $\pm$ 0.3 & $10^2$ & $10^2$ & 100 & $2$ & $4000$ & $0.005$ & $0.15$ & $20$ & $500$\\ 
		\cline{1-12}
		Shuttle & 99.8 $\pm$ 0.1 & 99.9 $\pm$ 0.1 & $10^5$ & $10^4$ & 100 & $2$ & $4000$ & $0.005$ & $0.05$ & $20$ & $500$ \\ 
		\cline{1-12}
		MNIST  & 95.7 $\pm$ 0.1 & 98.0 $\pm$ 0.1 & $1$ & $10^5$ & 100 & 2 & 4000 & $0.005$ & $0.15$ & 20 & $500$ \\ 
		\cline{1-12}
		CIFAR-10 & 47.3 $\pm$ 0.2 & 51.4 $\pm$ 0.2 & $10^8$ & $10^3$ & 100 & 2 & 4000 & 0.005 & 0.15 & 20 & 500 \\ 
		%		\cline{1-13}
		%		CIFAR-100 & 23.0 $\pm$ 0.2 & 23.9 $\pm$ 0.1 & 22.1 $\pm$ 0.1 & $10^8$ & $1$ & 100 & 2 & 1000 & 0.005 & 0.15 & 100 & 100 & \\ 
		\hline
	\end{tabular}
	%\vspace{-0.3cm}
\end{table*}

\subsubsection{Performance comparison, failure, mitigation}

So far we explored SSFN as a sequential learning in a forward manner. Our next experiment considers backpropagation for further improvement of SSFN, and compare with state-of-the-art performances. The results are shown in Table~\ref{table:PerformanceComparisonWithStateOfArt}. Here `state-of-the-art' performances are quoted from the literature, and references are mentioned in the table. We did not simulate them assuming they are reproducible. For the CNN, we perform simulations and testing for three datasets - Caltech101, MNIST, and CIFAR-10. These three datasets have image signals that are suitable in size for CNN implementation. The CNN code uses the following consecutive steps: 2D convolution, ReLU, 2D convolution, ReLU, 2D max-pooling, dropout, 2D convolution, ReLU, 2D convolution, ReLU, 2D max-pooling, dropout, dense, ReLU, dropout, dense, and finally softmax. The CNN signal flow structure is same for the three datasets. Parameters of the CNN for each of the three datasets are learned using RMS prop with 30 epochs\footnote{RMSprop is an optimization method available at: \url{http://www.cs.toronto.edu/~tijmen/csc321}}. In the Table~\ref{table:PerformanceComparisonWithStateOfArt}, we consider back propagation for optimization of weight matrices in SSFN that explores dependence between successive layers of SSFN. This is referred to as backpropagation optimized SSFN (bSSFN), discussed in the last paragraph of section~\ref{subsec:PLN algorithm_and_further_optimization}. Weight matrices of SSFN are used as initialization in bSSFN. 
%We did not use any constraint on weight matrices at the time of backpropagation and hence the bSSFN does not hold any specific structure in its weight matrices. 
The learning rate of gradient search in back propagation is chosen using cross validation. Backpropagation requires a significantly high computational resource, hours in our multi-processor server. Backpropagation for many instances of SSFN requires a considerable simulation time. Therefore we show result for one instance of SSFN for every dataset and the corresponding instance of bSSFN in the table. 
%The instance of PLN corresponds to the instance of bPLN. 
We do not try to show performance for the good instances among 50 Monte-Carlo simulations in the table as this is subjected to a careful selection procedure. %The rPLN result is shown as the average of Monte Carlo simulations. 
%Backpropagation leads to a tangible improvement compared to SSFN when the number of layers of the SSFN is rather low, for example, in the case of NORB dataset. A potential explanation of this behavior may be that large neural networks suffer significantly from a vanishing gradient problem. One way to mitigate this problem would be to incorporate a residual network structure inside the SSFN; this we did not explore in our work. 
%We mention that SSFN has a low complexity, requiring training time around order of ten minutes in the laptop. On the other hand, backpropagation requires hours in our server with multi-processors. 
We do not consider hSSFN and further optimization of hSSFN using backpropagation in the Table~\ref{table:PerformanceComparisonWithStateOfArt} as hand-tuning remains as an art. 
%Note that hSSFN performance is significantly better than SSFN for some datasets, for example, MNIST (see Table~\ref{table:tuned_PLN_performance_classification_saikat}). 
%It  Use of hSSFN followed by backpropagation could have given a better competitive edge than bSSFN as bSSFN used SSFN as its initialization.
% We include rPLN in the table due to its best performance of $99.7\%$ test accuracy for CIFAR-10 dataset.
%The rPLN performance for CIFAR-10 dataset is $99.7\%$; this performance is the best result achieved so far in comparison with state-of-the-art result $98.52\%$ \cite{Cubuk_AutoAugment_2018}. Note that rPLN does not do better than PLN for all datasets, for example, Caltech101 dataset. 
%CIFAR-100 still remains as a poor performance case.
In Table~\ref{table:PerformanceComparisonWithStateOfArt}, comparing with state-of-the-art, we find that SSFN and bSSFN provide a reasonable performance for seven datasets, but they are unable to compete with state-of-the-art for CIFAR-10 dataset. The SSFN fails for CIFAR-10.

It remains a question why the SSFN fails for CIFAR-10 dataset! We are yet to understand the relation between statistics of a dataset and performance of SSFN. The recent work \cite{Recht_CIFAR10_Generalization_2018} shows that it is non-trivial to achieve generalization in performance for CIFAR-10 dataset and high accuracy previously reported in literature may be questionable. At this point, we mention that we did not use various preprocessing methods on a training dataset for performance improvement. For example, the work of \cite{Cubuk_AutoAugment_2018} uses efficient data augmentation methods that we do not use. 
%Investigation using data augmentation is out of scope in this work. 
%We did not use any data augmentation method for performance improvement as our main interest is restricted to estimating size of the feedforward neural network SSFN. Investigation on data augmentation is out of scope of this work. 
Finally we comment on SSFN performance in comparison with CNN. The CNN signal flow structure is found to be good for the MNIST and CIFAR-10 datasets. The same CNN signal flow structure is not found very competitive for Caltech101 dataset. For a dataset, CNN requires hand tuning for its signal flow structure design followed by size selection. Structure selection for CNN considers appropriate use and judicious juxtaposition of convolutional layer, ReLU function, max pooling layer, fully connected layer, softmax, etc. Note that, in the case of Caltech101 dataset, we have used the 3000-dimensional feature vectors suggested in \cite{Label_Consistent_KSVD_2013} for SSFN and bSSFN; the CNN directly uses image signals with appropriate downsampling to $128 \times 128$ pixel size. The results for Caltech101 dataset show the importance of feature design using domain knowledge. 
%This is a noteworthy outcome of our experiment. 
On the other hand, the success of CNN for MNIST and CIFAR-10 datasets can be partially attributed to the convolutional structure in weight matrices as linear transform, max pooling operation in nonlinear transform design and dropout for regularization. 
A future work for improvement of SSFN is to explore the application of different signal transforms used in CNN. For example, we may explore use of convolutional filter. A convolution filter is associated with a circulant matrix. Therefore, to construct $\mathbf{W}_l$, we may explore in future use of (structured) circulant matrix instead of (unstructured) random instance based matrix $\mathbf{R}_l$ (see \eqref{eq:StructureOfWeightMatrix_LthLayer}).

\begin{table}[t]
	\centering
	\caption{Comparison of classification performances for SSFN, back propagation based SSFN (bSSFN), state-of-the-art methods in the literature and CNN}
	\vspace{-2mm}
	\label{table:PerformanceComparisonWithStateOfArt}
	\setlength{\tabcolsep}{4pt}
	\renewcommand{\arraystretch}{1.2}
	\begin{tabular}{ |c|c|c|c|c|c|} 
		\hline
		Dataset & {\begin{tabular}{@{}c@{}}SSFN \\ (one instance) \end{tabular}}  & {\begin{tabular}{@{}c@{}} bSSFN \\ (one instance) \end{tabular}}   &  
		{\begin{tabular}{@{}c@{}} state-of-the-art \\ (reference) \end{tabular}} & CNN \\  
		\hline \hline 
		Vowel & 61.17 & 61.17 & 64.94 \cite{Tang_HELM_2016} & - \\ 
		\hline
		%		Extended YaleB & 98.29 & \textbf{98.33} & 97.20 & \cite{Jiang_LCKSVD_2013}\\ 
		%		\hline
		%		AR & 98.25 & \textbf{98.28} & 97.80 & \cite{Jiang_LCKSVD_2013}\\ 
		%		\hline
		Satimage & 89.92 & 90.08 & 90.90 \cite{Jiang_LCKSVD_2013} & - \\ 
		\hline
		%		Scene15 & 99.34 & \textbf{99.35} & 92.90 & \cite{Jiang_LCKSVD_2013}\\ 
		%		\hline
		Caltech101 & 75.30 & 75.33 & 78.50 \cite{Kanan_RobustClassification_2010} & 45.77 \\ 
		\hline
		Letter & 95.52 & 95.65 & 95.82 \cite{Tang_HELM_2016} & - \\ 
		\hline
		NORB & 85.81 & 88.71 & 89.20 \cite{Salakhutdinov_DBM_2009} & - \\ 
		\hline
		Shuttle & 99.82 & 99.90 & 99.91 \cite{Huang_ELM_2012} & - \\ 
		\hline
		MNIST & 95.55 & 97.61 & 99.79 \cite{Wan_DropConnect_2013} & 99.33 \\ 
		\hline
		CIFAR-10 & 47.21 & 49.85 & 98.52 \cite{Cubuk_AutoAugment_2018} & 75.34 \\ 
		\hline
		%		CIFAR-100 & 22.95 & 23.51 & 22.1 & 84.80 \cite{DeVries_Cutout_2017}\\ 
		%		\hline
	\end{tabular}
\end{table}

\medmuskip=-2mu
\begin{figure*}[t!]
	\centering
	\def\svgwidth{\linewidth}
	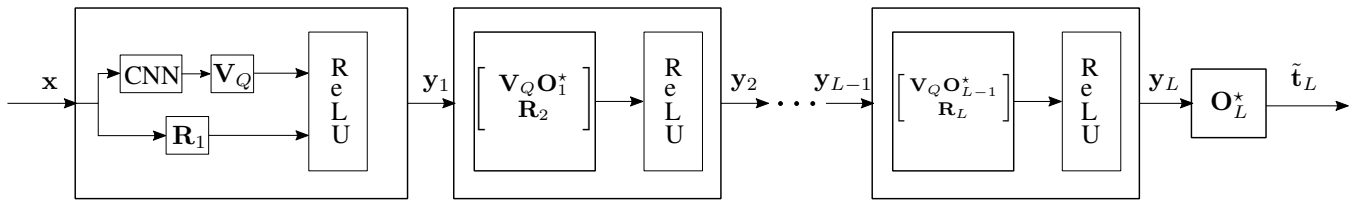	
	\caption{A hybrid system - the architecture of a multi-layer SSFN with $L$ layers where CNN output is used in the first layer.}
	\label{fig:CNN_Supported_MultiLayerPLN}
	%	\vspace{-12pt}
\end{figure*}
\medmuskip=4mu

We now design an ad-hoc approach to mitigate the limitation of SSFN for CIFAR-10 dataset. SSFN has a flexible system architecture that can use output of a successful method easily in its entry point. For example, we can use the output of CNN in SSFN. Note that we used least-squares output $\tilde{\mathbf{t}}_0^{\star} = \mathbf{O}_0^{\star} \mathbf{x}$ in the first layer of SSFN. Instead of the least-squares output, we can easily use CNN output in the first layer as a simple replacement. This leads to a hybrid system. The resulting hybrid system is shown in Figure~\ref{fig:CNN_Supported_MultiLayerPLN} which provides $76.4\%$ accuracy for the CIFAR-10 dataset. This result is $1.1\%$ better than CNN performance shown in Table~\ref{table:PerformanceComparisonWithStateOfArt}. Note that we have used the same set of hyperparameters as in Table \ref{table:Classification_performance_without_Zscore_saikat} for the SSFN part of the hybrid system. We have not tuned any additional parameter of the SSFN part.
We perhaps might further improve the result $76.4\%$ by hand-tuning the parameters of the system. This hand-tuning route is not exercised due to the aspect `limited need for human intervention'.

\section{Conclusions and Questions}

We conclude that it is possible to engineer an algorithm such that a feed-forward neural network can estimate its own size in a computationally efficient manner without a significant human involvement. A judicious combination of layer-wise learning approach, convex optimization and random matrix usage is useful. The method turns out to be resilient against variation in parameter tuning. In fact many parameters can remain same across datasets and tasks (see Table \ref{table:Classification_performance_without_Zscore_saikat}). Following our experimental results, we envisage that underlying statistics of a training dataset is a key factor for the SSFN size estimation. Size estimation and classification performance both are found consistent across Monte-Carlo simulations.  

%The proposed layer-wise sequential learning strategy helps to utilize advantage of convex optimization in addition with analytical form of regularization coefficients.  
%Classification performance of the proposed self size-estimating feed-forward network (SSFN) is reasonable compared to the state-of-the-art results for the seven datasets among eight datasets that we tested. The size of SSFN varies significantly across the eight datasets. The (unknown) statistics of a training dataset has an important role for estimating the size of SSFN.

We have observed competitive performance as well as failure compared to the state-of-the-art results. Currently we do not have theoretically motivated understanding of several questions: Why do we get consistent size and consistent performance across independent simulations? Why do we get the the failure case, or what is the data statistics that the method will fail? What is the main limitation in the SSFN system architecture? How to identify the limitation and develop a mitigation approach? These questions remain for future study.

\ifCLASSOPTIONcaptionsoff
  \newpage
\fi



%\bibliographystyle{IEEEtran}
%\begin{thebibliography}{1}

\bibliographystyle{IEEEtran}
\bibliography{ref,ref_alireza,biblio_saikat_ANN,biblio_saikat_CS1,biblio_saikat_Pub}

\begin{thebibliography}{10}
\providecommand{\url}[1]{#1}
\csname url@samestyle\endcsname
\providecommand{\newblock}{\relax}
\providecommand{\bibinfo}[2]{#2}
\providecommand{\BIBentrySTDinterwordspacing}{\spaceskip=0pt\relax}
\providecommand{\BIBentryALTinterwordstretchfactor}{4}
\providecommand{\BIBentryALTinterwordspacing}{\spaceskip=\fontdimen2\font plus
\BIBentryALTinterwordstretchfactor\fontdimen3\font minus
  \fontdimen4\font\relax}
\providecommand{\BIBforeignlanguage}[2]{{%
\expandafter\ifx\csname l@#1\endcsname\relax
\typeout{** WARNING: IEEEtran.bst: No hyphenation pattern has been}%
\typeout{** loaded for the language `#1'. Using the pattern for}%
\typeout{** the default language instead.}%
\else
\language=\csname l@#1\endcsname
\fi
#2}}
\providecommand{\BIBdecl}{\relax}
\BIBdecl

\bibitem{Connectionist_Handwriting_Recognition_2009}
A.~Graves, M.~Liwicki, S.~Fernández, R.~Bertolami, H.~Bunke, and
  J.~Schmidhuber, ``A novel connectionist system for unconstrained handwriting
  recognition,'' \emph{IEEE Transactions on Pattern Analysis and Machine
  Intelligence}, vol.~31, no.~5, pp. 855--868, May 2009.

\bibitem{Representation_Learning_Review_2013}
Y.~Bengio, A.~Courville, and P.~Vincent, ``Representation learning: A review
  and new perspectives,'' \emph{IEEE Transactions on Pattern Analysis and
  Machine Intelligence}, vol.~35, no.~8, pp. 1798--1828, Aug 2013.

\bibitem{3D_CNN_2013}
S.~Ji, W.~Xu, M.~Yang, and K.~Yu, ``3d convolutional neural networks for human
  action recognition,'' \emph{IEEE Transactions on Pattern Analysis and Machine
  Intelligence}, vol.~35, no.~1, pp. 221--231, Jan 2013.

\bibitem{Feedforward_LSTM_LanguageModeling_2015}
M.~Sundermeyer, H.~Ney, and R.~Schlüter, ``From feedforward to recurrent lstm
  neural networks for language modeling,'' \emph{IEEE/ACM Transactions on
  Audio, Speech, and Language Processing}, vol.~23, no.~3, pp. 517--529, March
  2015.

\bibitem{Bebis_FeedForwardNeuralNet_1994}
G.~Bebis and M.~Georgiopoulos, ``Feed-forward neural networks,'' \emph{IEEE
  Potentials}, vol.~13, no.~4, pp. 27--31, Oct 1994.

\bibitem{SizeMattersInANN_1999}
D.~Ellis and N.~Morgan, ``Size matters: an empirical study of neural network
  training for large vocabulary continuous speech recognition,'' in \emph{1999
  IEEE International Conference on Acoustics, Speech, and Signal Processing.},
  vol.~2, Mar 1999, pp. 1013--1016.

\bibitem{Cubuk_AutoAugment_2018}
E.~D. Cubuk, B.~Zoph, D.~Mane, V.~Vasudevan, and Q.~V. Le, ``Autoaugment:
  Learning augmentation policies from data,'' \emph{ar{X}iv preprint}, 2018.

\bibitem{BoydPCPE11}
S.~Boyd, N.~Parikh, E.~Chu, B.~Peleato, and J.~Eckstein, ``Distributed
  optimization and statistical learning via the alternating direction method of
  multipliers,'' \emph{Foundations and Trends in Machine Learning}, vol.~3,
  no.~1, pp. 1--122, 2011.

\bibitem{PLN_Saikat}
S.~Chatterjee, A.~M. Javid, M.~Sadeghi, P.~P. Mitra, and M.~Skoglund,
  ``Progressive learning for systematic design of large neural networks,''
  \emph{ar{X}iv preprint}, 2017.

\bibitem{Cascade-correlation_ANN_1990}
S.~E. Fahlman and C.~Lebiere, \emph{The cascade-correlation learning
  architecture}.\hskip 1em plus 0.5em minus 0.4em\relax Advances in neural
  information processing systems 2, 1990.

\bibitem{Training_MLPs_layer_by_layer_1996}
R.~Lengellé and T.~Denœux, ``Training mlps layer by layer using an objective
  function for internal representations,'' \emph{Neural Networks}, vol.~9,
  no.~1, pp. 83 -- 97, 1996.

\bibitem{lbdl15}
Y.~LeCun, Y.~Bengio, and G.~Hinton, ``Deep learning,'' \emph{Nature}, vol. 521,
  no. 7553, pp. 436--444, 2015.

\bibitem{blda09}
Y.~Bengio \emph{et~al.}, ``Learning deep architectures for {AI},''
  \emph{Foundations and trends{\textregistered} in Machine Learning}, vol.~2,
  no.~1, pp. --127, 2009.

\bibitem{GoodBC16}
I.~Goodfellow, Y.~Bengio, and A.~Courville, \emph{Deep Learning}.\hskip 1em
  plus 0.5em minus 0.4em\relax MIT Press, 2016.

\bibitem{Hinton_DeepCNN_2012}
A.~Krizhevsky, I.~Sutskever, and G.~E. Hinton, ``Imagenet classification with
  deep convolutional neural networks,'' in \emph{Advances in Neural Information
  Processing Systems 25}, 2012, pp. 1097--1105.

\bibitem{DeepResidualLearning_CVPR_2016}
K.~He, X.~Zhang, S.~Ren, and J.~Sun, ``Deep residual learning for image
  recognition,'' in \emph{2016 IEEE Conference on Computer Vision and Pattern
  Recognition (CVPR)}, June 2016, pp. 770--778.

\bibitem{RNN_based_language_model_2012}
T.~Mikolov, M.~Karafiát, L.~Burget, J.~Cernocký, and S.~Khudanpur,
  ``Recurrent neural network based language model,'' in \emph{INTERSPEECH
  2010}, vol.~2, 01 2010, pp. 1045--1048.

\bibitem{DBN_Hinton_2006}
G.~Hinton, S.~Osindero, and Y.~Teh, ``A fast learning algorithm for deep belief
  nets,'' \emph{Neural Computation}, vol.~18, no.~7, pp. 1527--1554, 2006.

\bibitem{Bengio_GreedyLayerWise_DNN_2007}
Y.~Bengio, P.~Lamblin, D.~Popovici, and H.~Larochelle, ``Greedy layer-wise
  training of deep networks,'' in \emph{Advances in Neural Information
  Processing Systems 19}.\hskip 1em plus 0.5em minus 0.4em\relax MIT Press,
  2007, pp. 153--160.

\bibitem{Ivakhnenko_Polynomial_Theory_of_Complex_Systems_1971}
A.~G. Ivakhnenko, ``Polynomial theory of complex systems,'' \emph{IEEE
  Transactions on Systems, Man, and Cybernetics}, vol. SMC-1, no.~4, pp.
  364--378, 1971.

\bibitem{blglw07}
Y.~Bengio, P.~Lamblin, D.~Popovici, H.~Larochelle \emph{et~al.}, ``Greedy
  layer-wise training of deep networks,'' \emph{Advances in neural information
  processing systems}, vol.~19, p. 153, 2007.

\bibitem{KulkK17}
M.~Kulkarni and S.~Karande, ``Layer-wise training of deep networks using kernel
  similarity,'' \emph{ar{X}iv preprint}, 2017.

\bibitem{HettCEHJW17}
C.~Hettinger, T.~Christensen, B.~Ehlert, J.~Humpherys, T.~Jarvis, and S.~Wade,
  ``Forward thinking: Building and training neural networks one layer at a
  time,'' \emph{ar{X}iv preprint}, 2017.

\bibitem{Larsen_regularized_ANN_1994}
J.~Larsen and L.~K. Hansen, ``Generalization performance of regularized neural
  network models,'' in \emph{Proceedings of IEEE Workshop on Neural Networks
  for Signal Processing}, Sep 1994, pp. 42--51.

\bibitem{Hinton_SoftWeight_1992}
S.~J. Nowlan and G.~E. Hinton, ``Simplifying neural networks by soft
  weight-sharing,'' \emph{Neural Computation}, vol.~4, no.~4, pp. 473--493,
  July 1992.

\bibitem{Hinton_Dropout_2014}
N.~Srivastava, G.~Hinton, A.~Krizhevsky, I.~Sutskever, and R.~Salakhutdinov,
  ``Dropout: A simple way to prevent neural networks from overfitting,''
  \emph{Journal of Machine Learning Research}, vol.~15, pp. 1929--1958, 2014.

\bibitem{HuanZS06}
G.-B. Huang, Q.-Y. Zhu, and C.-K. Siew, ``Extreme learning machine: {T}heory
  and applications,'' \emph{Neurocomputing}, vol.~70, no. 1--3, pp. 489--501,
  2006.

\bibitem{Huang_What_are_ELM_2015}
G.-B. Huang, ``What are extreme learning machines? filling the gap between
  frank rosenblatt's dream and john von neumann's puzzle,'' \emph{Cognitive
  Computation}, vol.~7, no.~3, pp. 263--278, Jun 2015.

\bibitem{Huang_ELM_2012}
G.~Huang, H.~Zhou, X.~Ding, and R.~Zhang, ``Extreme learning machine for
  regression and multiclass classification,'' \emph{IEEE Transactions on
  Systems, Man, and Cybernetics}, vol.~42, no.~2, pp. 513--529, 2012.

\bibitem{Review_ANN_random_weights_2018}
W.~Cao, X.~Wanga, Z.~Minga, and J.~Gao, ``A review on neural networks with
  random weights,'' \emph{Neurocomputing}, vol. 275, pp. 278--287, 2018.

\bibitem{Schimdt_NeuralNetWithRandomWeights_1992}
W.~Schmidt, M.~Kraaijveld, and R.~Duin, ``Feed forward neural networks with
  random weights,'' in \emph{Proc. 11th IAPR Int Conf Vol. II, B: Pattern
  Recognition Methodology and Systems}, 1992.

\bibitem{PAO_NN_RandomVectors_1994}
Y.-H. Pao, G.-H. Park, and D.~J. Sobajic, ``Learning and generalization
  characteristics of the random vector functional-link net,''
  \emph{Neurocomputing}, vol.~6, no.~2, pp. 163 -- 180, 1994.

\bibitem{Igelnik_FunctionApproximation_StochasticChoice_1995}
B.~Igelnik and Y.-H. Pao, ``Stochastic choice of basis functions in adaptive
  function approximation and the functional-link net,'' \emph{IEEE Transactions
  on Neural Networks}, vol.~6, no.~6, pp. 1320--1329, 1995.

\bibitem{LU_ANN_RandomWeights_2014}
J.~Lu, J.~Zhao, and F.~Cao, ``Extended feed forward neural networks with random
  weights for face recognition,'' \emph{Neurocomputing}, vol. 136, pp. 96 --
  102, 2014.

\bibitem{Rahimi_RandomKitchenSinks_NIPS_2008}
A.~Rahimi and B.~Recht, ``Weighted sums of random kitchen sinks: Replacing
  minimization with randomization in learning,'' in \emph{Advances in Neural
  Information Processing Systems 21}, 2009, pp. 1313--1320.

\bibitem{FastFood_AlexSmola_2013}
Q.~Le, T.~Sarlos, and A.~Smola, ``Fastfood - approximating kernel expansions in
  loglinear time,'' in \emph{30th International Conference on Machine Learning
  (ICML)}, 2013.

\bibitem{Donoho_2006_Compressed_sensing}
D.~Donoho, ``Compressed sensing,'' \emph{Information Theory, IEEE Transactions
  on}, vol.~52, no.~4, pp. 1289 --1306, april 2006.

\bibitem{CS_introduction_Candes_Wakin_2008}
E.~Candes and M.~Wakin, ``An introduction to compressive sampling,'' \emph{IEEE
  Signal Proc. Magazine}, vol.~25, pp. 21--30, march 2008.

\bibitem{Vehkapera_Kabashima_Chatterjee_TIT_2016}
M.~Vehkaper\"{a}, Y.~Kabashima, and S.~Chatterjee, ``Analysis of regularized ls
  reconstruction and random matrix ensembles in compressed sensing,''
  \emph{IEEE Transactions on Information Theory}, vol.~62, no.~4, pp.
  2100--2124, April 2016.

\bibitem{Sastry_2009_Face_recognition}
J.~Wright, A.~Yang, A.~Ganesh, S.~Sastry, and M.~Yi, ``Robust face recognition
  via sparse representation,'' \emph{Pattern Analysis and Machine Intelligence,
  IEEE Transactions on}, vol.~31, no.~2, pp. 210 --227, 2009.

\bibitem{Label_Consistent_KSVD_2013}
Z.~Jiang, Z.~Lin, and L.~S. Davis, ``Label consistent k-svd: Learning a
  discriminative dictionary for recognition,'' \emph{IEEE Transactions on
  Pattern Analysis and Machine Intelligence}, vol.~35, no.~11, pp. 2651--2664,
  2013.

\bibitem{Learning_fast_approximations_of_sparse_coding_LeCun_2010}
K.~Gregor and Y.~LeCun, ``Learning fast approximations of sparse coding,'' in
  \emph{International Conference on Machine Learning}, 2010.

\bibitem{Learned_Convolutional_Sparse_Coding_2018}
H.~{Sreter} and R.~{Giryes}, ``Learned convolutional sparse coding,'' in
  \emph{International Conference on Acoustics, Speech and Signal Processing
  (ICASSP)}, 2018, pp. 2191--2195.

\bibitem{Algorithm_Unrolling_YCEldar_2019}
Y.~{Li}, M.~{Tofighi}, V.~{Monga}, and Y.~C. {Eldar}, ``An algorithm unrolling
  approach to deep image deblurring,'' in \emph{International Conference on
  Acoustics, Speech and Signal Processing (ICASSP)}, 2019, pp. 7675--7679.

\bibitem{Todd_1988}
P.~Todd, ``Evolutionary methods for connectionist architectures.''
  \emph{Psychology Dept., Stanford University, unpublished Manuscript}, 1988.

\bibitem{Miller_Todd_1989}
F.~Miller, P.~Todd, and S.~Hegde, ``Designing neural networks using genetic
  algorithms,'' in \emph{Proceedings of the third international conference on
  Genetic algorithms}, 1989, pp. 379--384.

\bibitem{Kitano_1990}
H.~Kitano, \emph{Designing neural networks using genetic algorithms with graph
  generation system}.\hskip 1em plus 0.5em minus 0.4em\relax Complex systems,
  1990.

\bibitem{NAS_RL_2017}
B.~Zoph and Q.~V. Le, ``Neural architecture search with reinforcement
  learning,'' in \emph{International Conference on Learning Representations},
  2017.

\bibitem{pmlr-v64-mendoza_towards_2016}
H.~Mendoza, A.~Klein, M.~Feurer, J.~T. Springenberg, and F.~Hutter, ``Towards
  automatically-tuned neural networks,'' in \emph{Proceedings of the Workshop
  on Automatic Machine Learning}, 2016, pp. 58--65.

\bibitem{Bayesian_NAS_2019}
G.~Dikov, P.~van~der Smagt, and J.~Bayer, ``Bayesian learning of neural network
  architectures,'' 2019.

\bibitem{NAS_Survey_2018}
T.~Elsken, J.~H. Metzen, and F.~Hutter, ``Neural architecture search: A
  survey,'' \emph{Journal of Machine Learning Research}, vol.~20, pp. 1--21,
  2019.

\bibitem{MaasHN13}
A.~L. Maas, A.~Y. Hannun, and A.~Y. Ng, ``Rectifier nonlinearities improve
  neural network acoustic models,'' in \emph{Proc. ICML}, vol.~30, 2013.

\bibitem{ADAM_2015}
D.~P. Kingma and J.~Ba, ``Adam: A method for stochastic optimization,'' in
  \emph{Proceedings of the 3rd International Conference on Learning
  Representations (ICLR)}, 2015.

\bibitem{Recht_CIFAR10_Generalization_2018}
B.~Recht, R.~Roelofs, L.~Schmidt, and V.~Shankar, ``Do cifar-10 classifiers
  generalize to cifar-10?'' \emph{ar{X}iv preprint}, 2018.

\bibitem{Tang_HELM_2016}
J.~Tang, C.~Deng, and G.~Huang, ``Extreme learning machine for multilayer
  perceptron,'' \emph{IEEE Transactions on Neural Networks and Learning
  Systems}, vol.~27, no.~4, pp. 809--821, April 2016.

\bibitem{Jiang_LCKSVD_2013}
Z.~Jiang, Z.~Lin, and L.~S. Davis, ``Label consistent k-svd: Learning a
  discriminative dictionary for recognition,'' \emph{IEEE Transactions on
  Pattern Analysis and Machine Intelligence}, vol.~35, no.~11, pp. 2651--2664,
  2013.

\bibitem{Kanan_RobustClassification_2010}
C.~Kanan and G.~Cottrell, ``Robust classification of objects, faces, and
  flowers using natural image statistics,'' in \emph{IEEE Computer Society
  Conference on Computer Vision and Pattern Recognition}, 2010.

\bibitem{Salakhutdinov_DBM_2009}
R.~Salakhutdinov and G.~Hinton, ``Deep boltzmann machines,'' in
  \emph{Proceedings of the Twelth International Conference on Artificial
  Intelligence and Statistics}.\hskip 1em plus 0.5em minus 0.4em\relax PMLR,
  2009.

\bibitem{Wan_DropConnect_2013}
L.~Wan, M.~Zeiler, S.~Zhang, Y.~L. Cun, and R.~Fergus, ``Regularization of
  neural networks using dropconnect,'' in \emph{Proceedings of the 30th
  International Conference on Machine Learning}.\hskip 1em plus 0.5em minus
  0.4em\relax PMLR, 2013.

\end{thebibliography}

%H.~Kopka and P.~W. Daly, \emph{A Guide to {\LaTeX}}, 3rd~ed.\hskip 1em plus

%\end{thebibliography}


%\end{IEEEbiography}


%\newpage



%\vfill


\end{document}